%% file: icml2026/main.tex
\documentclass{article}

\usepackage{microtype}
\usepackage{graphicx}
\usepackage{subcaption}
\usepackage{booktabs} 
\usepackage{tikz}
\usetikzlibrary{decorations.pathreplacing, arrows.meta, calc}

\usepackage{amsmath}
\usepackage{amssymb}
\usepackage{array}
\usepackage{mathtools}
\usepackage{amsthm}
\usepackage{thmtools} 
\usepackage{thm-restate}
\usepackage{enumitem}

\usepackage{hyperref}

\usepackage[preprint]{icml2026}

\usepackage[capitalize,noabbrev]{cleveref}

\newcommand{\calH}{\mathcal{H}}
\newcommand{\calD}{\mathcal{D}}
\newcommand{\calE}{\mathcal{E}}
\newcommand{\calF}{\mathcal{F}}
\newcommand{\calR}{\mathcal{R}}
\newcommand{\calG}{\mathcal{G}}
\newcommand{\calX}{\mathcal{X}}
\newcommand{\calY}{\mathcal{Y}}

\newcommand{\hatxi}{\hat{\xi}}
\newcommand{\hatu}{\hat{u}}
\newcommand{\estimate}{\hat{\theta}}
\newcommand{\est}{\text{def}}
\newcommand{\err}{\text{err}}

\newcommand{\event}{\mathcal{W}}

\newcommand{\E}{\mathop{\mathbb{E}}}

\DeclareMathOperator*{\argmax}{arg\,max}
\DeclareMathOperator*{\argmin}{arg\,min}
\newcommand\independent{\protect\mathpalette{\protect\independenT}{\perp}}
\def\independenT#1#2{\mathrel{\rlap{$#1#2$}\mkern2mu{#1#2}}}

\newcommand{\Unif}{\text{Unif}}

\newcommand{\con}{\text{con}}
\newcommand{\elim}{\text{elim}}
\newcommand{\deltaprior}{\delta_{\text{prior}}}

\newcommand{\propThreshold}{\beta_{\max}}

\newcommand{\hatpi}{\hat{\pi}}
\newcommand{\hatf}{\hat{f}}
\newcommand{\hatg}{\hat{g}}
\newcommand{\hatR}{\hat{R}}
\newcommand{\hatU}{\hat{U}}

\newcommand{\A}{\mathcal{A}}

\newcommand{\calA}{\mathcal{A}}

\newcommand{\BlackBoxRegressor}{\text{BlackBoxRegressor}}
\newcommand{\BlackBoxPolicyEvaluator}{\text{BlackBoxPolicyEvaluator}}
\newcommand{\CBErrorEstimator}{\text{CBErrorEstimator}}
\newcommand{\CBElimErrorEstimator}{\text{CBElimErrorEstimator}}
\newcommand{\CBConErrorEstimator}{\text{CBConErrorEstimator}}
\newcommand{\ArmEliminator}{\text{ArmEliminator}}

\newcommand{\width}{\text{width}}
\newcommand{\mean}{\text{mean}}
\newcommand{\ValidEliminator}{\text{ValidEliminator}}

\newcommand{\ValidateEliminator}{\text{ValidateEliminator}}
\newcommand{\IntervalMultiplier}{\gamma}
\newcommand{\BlackBoxCIEstimator}{\text{BlackBoxCIEstimator}}
\theoremstyle{plain}
\newtheorem{theorem}{Theorem}[section]

\newtheorem{lemma}[theorem]{Lemma}
\newtheorem{corollary}[theorem]{Corollary}
\theoremstyle{definition}

\newtheorem{assumption}[theorem]{Assumption}
\theoremstyle{remark}
\newtheorem{remark}[theorem]{Remark}

\usepackage{todonotes}

\icmltitlerunning{Error Estimation}

\begin{document}
\twocolumn[
  \icmltitle{Data-driven Error Estimation:\\ Excess Risk Bounds without Class Complexity as Input}



  \icmlsetsymbol{equal}{*}

  \begin{icmlauthorlist}
    \icmlauthor{Sanath Kumar Krishnmamurthy}{equal,meta}
    \icmlauthor{Anna Lyubarskaja}{equal,icme}
    \icmlauthor{Emma Brunskill}{cs}
    \icmlauthor{Susan Athey}{gsb,hai}
  \end{icmlauthorlist}

  \icmlaffiliation{icme}{Institute of Computational and Mathematical Engineering, Stanford University}
  \icmlaffiliation{meta}{Meta AI, Sunnyvale, CA}
  \icmlaffiliation{cs}{Department of Computer Science, Stanford University}
  \icmlaffiliation{gsb}{Graduate School of Business, Stanford University}
  \icmlaffiliation{hai}{Stanford Institute for Human-Centered AI}

  \icmlcorrespondingauthor{Anna Lyubarskaja}{annalyu@stanford.edu}
  \icmlcorrespondingauthor{Sanath Kumar Krishnamurthy}{sanathsk@meta.com}

  \icmlkeywords{Machine Learning, ICML}

  \vskip 0.3in
]


\printAffiliationsAndNotice{}  

\begin{abstract}
  Constructing confidence intervals that are simultaneously valid across a class of estimates is central to tasks such as multiple mean estimation, generalization guarantees, and adaptive experimental design. We frame this as an ``error estimation problem," where the goal is to determine a high-probability upper bound on the maximum error for a class of estimates. We propose an entirely data-driven approach that derives such bounds for both finite and infinite class settings, naturally adapting to a potentially unknown correlation structure of random errors. Notably, our method does not require class complexity as an input, overcoming a major limitation of existing approaches. We present our simple yet general solution and demonstrate applications to simultaneous confidence intervals, excess-risk control and optimizing exploration in contextual bandit algorithms.
\end{abstract}

\section{Introduction}
\label{sec:intro}
We introduce the following ``error estimation problem": For each $h$ in a class of estimation tasks $\calH$, we define a true estimand $\theta_h$ and its estimate $\estimate_h$. The error $e_h$, is a measure of distance between the estimand and the estimate. Our goal is to derive a high-probability upper bound on $\max_{h \in \calH}e_h$. 

\paragraph{Example 1: Subgroup Means} Consider an educational assessment where $\calH$ represents a (finite or infinite) set of various demographic or academic subgroups (such as `all students", ``male 4th graders", etc.). Setting $\theta_h$ to be the true mean test score for subgroup $h$, and $\estimate_h$ as its sample mean estimate, we may wish to construct simultaneously valid confidence intervals (CI) for all subgroup mean test scores. By providing an upper bound for $\max_{h \in \calH}e_h$, a solution to the error estimation problem directly yields such intervals.

\paragraph{Example 2: Generalization and Excess Risk} A second example, with an infinite class of tasks, arises in supervised learning. The excess risk \cite{shalev2014understanding} -- the gap between a learned model's error and the error of the best model in the model class --  is a critical input into sequential decision-making tasks, such as contextual bandits, where the bounds are needed to inform the tradeoff between exploration (gathering more data in order to identify better policies) and exploitation (taking advantage of a policy that is near-optimal) \cite{agarwal2019reinforcement}. We approach this by setting up an error estimation problem where $\calH$ represents a model class, $\theta_h$ is the true risk of a model $h \in \calH$, and $\estimate_h$ is the empirical risk over a dataset. Bounding $\max_{h \in \calH}e_h$ then gives us a generalization bound across $\calH$, which can be combined with localization procedures to control excess risk.

The literature has proposed several approaches to the error estimation problem. While single errors are often bounded using concentration inequalities or the central limit theorem (CLT), the problem of constructing simultaneous guarantees over an entire family of estimates is more challenging. For finite estimation classes (e.g. a finite number of subgroups, as in Example 1), a standard (but conservative) solution is the union bound, which adjusts individual error bounds based on the number of subgroups, $|\calH|$; practical empirical alternatives such as bootstrapping are widely used. Existing methods for infinite classes express uncertainty as a function of class complexity  (e.g. VC dimension or Rademacher complexity), which is often difficult to characterize, leading to the use of conservative estimates of complexity and thus overly conservative upper bounds and resulting CIs. Nonetheless, no widely adopted empirical alternatives exist. 

\paragraph{Contributions.}
(i) \textbf{A data-splitting wrapper for uniform guarantees (finite-sample under finite-sample pointwise bounds):} given any pointwise high-probability error bound that holds for each fixed $h$, we obtain a uniform bound on $\sup_{h\in\calH} e_h$. This applies to both finite and infinite classes and automatically adapts to correlations across tasks \footnote{Although these are uniform bounds, in our setup $e_h$ can be normalized for each $h$, allowing us to get simultaneously valid variance dependent bounds -- see \Cref{sec:ci-multiple-mean}}.
(ii) \textbf{A computable, data-dependent complexity proxy:}
in supervised learning, the maximization objective induced by our framework coincides with a standard symmetrization/discrepancy quantity and yields a practical analogue of Rademacher-style complexity (formal connections and tightness in Appendix~\ref{app:rad}).
(iii) \textbf{Data-driven localization:} when uniform error bounds are unnecessary, we use an iterative localization procedure that yields tighter bounds by progressively excluding models less relevant to the objective—for example, excluding high-risk models when bounding excess risk.
(iv) \textbf{Contextual bandits beyond realizability:}
we show how error estimation yields exploration sets and bonuses that can be plugged into modular contextual bandit pipelines while retaining theoretical guarantees.


Contextual bandit algorithms require a measure of estimation error/uncertainty to guide exploration. While existing algorithms (with tight theoretical guarantees) rely on mathematically derived class complexity bounds to quantify such errors \cite{foster2023foundations}, our error estimation approach can achieve tight error/uncertainty bounds without needing such a priori mathematical analysis. Building on insights from \cite{foster2020instance,krishnamurthy2023proportional, krishnamurthy2024towards}, we show how error estimation enables the development of theoretically sound contextual bandit algorithms that accommodate a wide range of policy classes.

\subsection{Related Work}
\label{sec:related_work}

Our approach to the error estimation problem can be contrasted with several alternatives. For the finite case, beyond the classical literature on multiple hypothesis testing \cite{efron_large-scale_2010}, \textbf{bootstrapping} is particularly effective in the case of large but finite classes with potentially correlated errors \citep[see e.g.][]{romano_exact_2005}. When statistical complexity is bounded by a polynomial function of sample size, bootstrap-based methods have been applied to construct asymptotically valid confidence intervals for the maximum ($L^{\infty}$) error \citep{gine1990bootstrapping,chernozhukov2013gaussian,chernozhukov2023high}. However, (1) its guarantees typically require finite-dimensional or high-dimensional regimes with additional regularity/complexity control, (2) it only provides asymptotic guarantees, and (3) the number of bootstrap samples, $B$ must be chosen heuristically and scales poorly with model complexity.

In the infinite case, simultaneous error bounds are typically derived by analyzing \textbf{Rademacher complexity} or \textbf{VC dimension} \citep{shalev2014understanding,koltchinskii2011oracle}. These bounds are difficult to analyze and often fail to explain generalization behavior in modern overparameterized settings \cite{zhang2017understanding}. When our framework is applied to the generalization setting from Example 2, the resulting objective recovers an empirical analogue of Rademacher complexity, with data splitting playing the role of a symmetrization trick in Rademacher-based proofs. This coincides with the \textit{discrepancy function} introduced in \cite{bartlett_model_2002}. However, our broader framework (i) holds under weaker distribution assumptions (allowing for applications to the contextual bandit setting), and (ii) is empirically validated in representative settings. We also show that the objective is upper-bounded by the true Rademacher complexity (see \Cref{app:rad}), ensuring both practical feasibility and theoretical guarantees. \Cref{sec:localizing} introduces a data-driven localization procedure, analogous to theoretical localized complexities \cite{koltchinskii2011oracle}, allowing us to tighten excess risk bounds with no additional mathematical analysis.  

 \textbf{PAC-Bayes bounds} \cite{alquier2021user} offer an alternative to classical complexity-based generalization bounds by incorporating a prior over the hypothesis class \cite{dziugaite_computing_2017}. While good priors, concentrated around effective models, can provide tighter generalization bounds, uninformative priors result in no advantage over traditional class complexity-based bounds. Specifying good priors is hard in practice \cite{dziugaite2017computing}. Our solution to error estimation simultaneously bounds every error of interest, and doesn't require an informative prior to achieve an improvement over traditional methods.

The statistical learning theory bounds described above are critical for theoretically sound algorithms for active learning, contextual bandits, and reinforcement learning (RL). For instance, exploration strategies typically rely on the complexity of the policy class as an explicit input to estimate uncertainty \citep{agarwal2014taming, foster2020instance}. Our framework replaces these bounds with empirically estimated quantities, enabling a data-driven estimation of the error. We anticipate that error estimation can similarly improve on the statistical learning theory bounds in active learning and RL. 

Our work is also related in spirit to \textbf{conformal prediction}, which uses data splitting to construct marginal prediction intervals with coverage around individual test points \cite{vovk_conformal_2005}. While both approaches share the underlying idea of reserving a portion of the data to assess uncertainty, the goals and guarantees differ. Conformal methods provide marginal coverage for predictions on new, unseen instances, while our framework provides uniform error bounds over a class of estimated parameters.

It is helpful to note that our work is not closely related to constructing anytime-valid CIs \citep[e.g.,][]{ramdas2023game}, as we do not explicitly model or consider time dependent structure in our data. 

\section{Error Estimation Theory}
\label{sec:eetheory}
In this section, we introduce fundamental notation for error estimation (\Cref{sec:preliminaries-for-error-estimation}), present the main idea (\Cref{sec:max-error-estimation}) and propose a localization procedure for refined bounds (\Cref{sec:localizing}).

\subsection{Problem Setup}
\label{sec:preliminaries-for-error-estimation}

We begin by formalizing the error estimation problem. We fix an index set $\calH$ that labels a (finite or infinite) family of estimation tasks (the set of subgroups in Example 1), and a random \textbf{defining dataset} $S_\est$. For each $h \in \calH$, we associate three quantities.
\begin{itemize}[noitemsep]
    \item $\theta_h$: an unknown \textbf{target estimand} (the true mean test score for subgroup $h$),
    \item $\estimate_h$: an \textbf{estimate} of $\theta_h$ estimated from $S_{\est}$ (sample mean test score for subgroup $h$),
    \item $e_h$: an unknown real-valued \textbf{error} measuring ``proximity" between $\estimate_h$ and $\theta_h$.
\end{itemize}

Let $\calE$ be a function that maps a realization $s$ of $S_\est$, to the class of estimand-estimate-error tuples associated with each estimation task/index. $\calE(s) :=\{(h,\estimate_h(s),\theta_h,e_h(s))|h\in\calH\}$.\footnote{%
In some applications $\calH$ or $\theta_h$ may themselves depend on $S_{\est}$; see Sections \ref{sec:excess-risk-error-estimation}, \ref{sec:multiple-hyps}, and \ref{sec:cberrorestimation}.} To lighten the notation, we refer to $e_h(s)$ and $\estimate_h(s)$ as simply $e_h$ and $\estimate_h$ where the dependence on $s$ is clear from context. For a given confidence parameter $\delta\in(0,1)$, the goal of error estimation is to estimate a high-probability (holds with probability at least $1-\delta$) upper bound on the maximum error ($\max_{h\in\calH}e_h$).

To do so, we introduce $S_{\err}$, an independent \textbf{error estimation dataset}, drawn from some distribution $\calD$, that may depend on the realized $S_{\est}=s$.\footnote{For example, in reinforcement learning, future sampling distributions may depend on earlier estimates. However, in other estimation settings, estimating some parameters of a given distribution would not affect the distribution of future/holdout samples.}  Once $\calD$ is defined, $S_\err$ and its realizations do not further depend on $s$. We often enforce such conditions by ensuring that the random dataset $S_{\err}$ is either a newly sampled dataset from $\calD$ or a holdout dataset that wasn't used to construct any of the estimates $\estimate_h (s)$ in $\calE(s)$. 

Bounding a single error $e_h$ is often straightforward using classical concentration inequalities or asymptotic results. Bounding the maximum, $\max_{h \in \calH}e_h$, on the other hand, is more statistically challenging. A common approach is to build on single error bounds to construct a uniform guarantee. For instance, the union bound applies a correction to single error confidence levels to ensure simultaneous validity. Similarly, our solution to the error estimation problem assumes access to a single high-probability estimator (see \Cref{ass:base-error-estimate}). Our main insight is that this problem can be reduced to \Cref{thm:max-error-estimation-result}, which constructs a high-probability upper bound on the maximum error (that is, $\max_{h \in \calH}e_h$)  -- without needing bounds on the complexity or size of the class $\calH$ as an input. 

\begin{assumption}[Upper Bounds for Individual Errors]
\label{ass:base-error-estimate}
    We have access to error upper bound estimators $\hatu: (S_\err \times \mathcal{H} \times (0,1)) \to \mathbb{R}$ that satisfy the following condition: for all $h\in\calH$ and $\lambda\in(0,1)$,  
    \[\Pr_{S_{\err}\sim\calD}(\hatu(S_{\err},h,\lambda) \ge e_h|\calE(s))\geq  1 - \lambda.\]
    
\end{assumption}

\subsection{Main Insight: Data-Driven Error Estimation}
\label{sec:max-error-estimation}

We now formalize the key technical insight: Uniform control over $\max_{h \in \calH}e_h$ reduces to constructing  pointwise $(1-\delta)$ bound that holds for every fixed $h$. Since the maximizing index $\tilde h \in \arg\max_{h\in\calH} e_h$ is measurable with respect to $S_{\est}$, conditioning makes it fixed with respect to the randomness in $S_{\err}$, yielding a uniform guarantee without an explicit complexity term.

\begin{theorem}[Data-Driven Upper Bound on Max Error]
\label{thm:max-error-estimation-result}
    Suppose \Cref{ass:base-error-estimate} holds. Let $\hatxi$ denote our estimated upper bound on the maximum error, $\hatxi(S_{\err},\delta) := \max_{h\in\calH}\hatu(S_{\err}, h,\delta)$. Then for any $\delta\in(0,1)$, we have that \eqref{eq:max-error-guarantee} and \eqref{eq:max-error-guarantee-wo-conditioning} hold.\footnote{Neither the definition of $\hatxi$ or \eqref{eq:max-error-guarantee} (main guarantee) explicitly depend on the complexity size of class $\calH$.} \footnote{When $\calH$ is an infinite class, replace $\max_{h\in \calH}$ by $\sup_{h \in \calH}$. See \Cref{sec:local_proof} for details}
    \begin{equation}
    \label{eq:max-error-guarantee}
        \Pr_{S_{\err}\sim\calD}\Big(\hatxi(S_{\err},\delta) \geq \max_{h\in\calH}e_h\Big|\calE(s)\Big)\geq 1-\delta
    \end{equation}
    \begin{equation}
    \label{eq:max-error-guarantee-wo-conditioning}
    \Pr_{(S_{\est},S_{\err})}\Big(\hatxi(S_{\err},\delta) \geq \max_{h\in\calH}e_h\Big)\geq 1-\delta
    \end{equation}
\end{theorem}

\begin{proof}
    Let the index $\Tilde{h}\in\arg\max_{h\in\calH}e_h$ and the event $\event (S):=\{\hatu(S,\Tilde{h},\delta) \geq e_{\Tilde{h}}\}$. We first show that $\event(S_{\err})$ implies that $\hatxi(S_{\err},\delta)$ upper bounds the maximum error. Under $\event(S_{\err})$, we have $e_{\Tilde{h}} = \max_{h \in \calH}e_h$ and 
    \begin{equation}
    \label{eq:main-argument-in-hp-event}
    \begin{aligned}
    e_{\Tilde{h}} \stackrel{(i)}{\leq} \hatu(S_{\err},\Tilde{h},\delta) \stackrel{(ii)}{\leq} \max_{h\in\calH}\hatu(S_{\err},h,\delta)  &\stackrel{(iii)}{=:} \hatxi(S_{\err},\delta). 
    \end{aligned}
    \end{equation}
    Here (i) follows from $\event(S_{\err})$, (ii) follows from $\Tilde{h}\in\calH$, and (iii) follows from the definition of $\hatxi$. We now utilize \eqref{eq:main-argument-in-hp-event} and \Cref{ass:base-error-estimate} to show that \eqref{eq:max-error-guarantee} holds.
\begin{equation}
\label{eq:final-probability-argument}
\begin{aligned}
\Pr_{S{\err}\sim\calD}&\Big(\hatxi(S_{\err},\delta) \geq \max_{h\in\calH}e_h\Big|\calE(s)\Big)\\ & \stackrel{(i)}{\ge} \Pr_{S_{\err}\sim\calD}\Big(\hatu(S_{\err},\Tilde{h},\delta) \geq e_{\Tilde{h}}\Big|\calE(s)\Big) & \stackrel{(ii)}{\geq} 1-\delta
\end{aligned}
\end{equation}
    Here (i) follows from \eqref{eq:main-argument-in-hp-event} and the definition of $\event$, and (ii) follows from \Cref{ass:base-error-estimate} -- showing that \eqref{eq:max-error-guarantee} holds. Now by the law of total expectation\footnote{If $I = \mathbf{1}_{\{\hatxi(S_{\err},\delta) \geq \max_{h\in\calH}e_h\}}$, the law of total expectation states that $\mathbb{E}_{S_\est}[\mathbb{E}_{S_\err}[I|\calE(s)]] = \mathbb{E}_{S_\est, S_\err}[I]$} and \eqref{eq:max-error-guarantee}, we also have that \eqref{eq:max-error-guarantee-wo-conditioning} holds.
\end{proof}
 
\Cref{thm:max-error-estimation-result} is a general reduction: pointwise finite-sample error bounds on $S_\err$ immediately yield uniform guarantees over $S_\est$-dependent classes. In contrast to a union bound approach, the definition of $\hatxi(S_{\err},\delta)$ does not explicitly depend on the complexity or size of class $\calH$. Note that the dependence on $\calH$ occurs implicitly in the maximization over the class $\calH$. For finite classes, this maximization is straightforward. For infinite classes, our guarantee is stated for $\sup_{h\in\calH}\hatu(S_{\err},h,\delta)$; in practice we approximate this supremum with optimization heuristics (e.g., SGD with multiple restarts).

\subsection{Localizing error bounds}

\label{sec:localizing}
Traditional complexity bounds are frequently enhanced by employing a localization argument by bounding over a subclass near the optimum \cite{koltchinskii2011oracle}. We apply a similar  argument to data-driven bounds. Suppose that rather than bounding $\max_h e_h$, we are interested in an upper bound for the error $e_{h^*}$ for the instance $h^* :=\argmax_h \theta_h$ \footnote{In fact $h^*$ can be any fixed instance (possibly depending on $\calE$) independent of $S_\err$. We focus on $h^* :=\argmax_h \theta_h$ for clarity, other choices follow similar reasoning}. In this case, rather than having to maximize an upper bound across the whole class $\calH$, it is enough to maximize across any class that contains $h^*$. 

To construct such classes, we introduce \Cref{ass:h*bound} which posits the existence of a lower bound on $\theta_{h^*}$. This assumption arises naturally in settings where problem-specific characteristics provide prior knowledge.\footnote{For instance, in \Cref{sec:excess-risk-error-estimation}, excess risk is the maximum difference between the risk of the given model and the risk of any other model in the class -- hence fits the localization setup of \Cref{ass:h*bound} with $c=0$. }

\begin{assumption}[Lower Bound for Given Instance]
\label{ass:h*bound}
    We have an a priori lower bound on the maximum estimand. Specifically, there exists some constant $c$ such that $c \le \max_h \theta_{h}$.
\end{assumption}

In this setting, we base our result on a weaker alternative to \Cref{ass:base-error-estimate} (\Cref{ass:specific-instance-error-estimate}). 
\begin{assumption}[Error Estimation for Specific Instance]
\label{ass:specific-instance-error-estimate}
    We have access to error upper bound estimators $\hatu: (S_\err \times \mathcal{H} \times (0,1)) \to \mathbb{R}$ which satisfy
    \[\Pr_{S_{\err} \sim \calD}(e_{h^*} \le \hatu(S_{\err}, h^*, \delta)|\calE(s)) \ge 1-\delta\] for $h^* = \argmax_h \theta_h$, and any $\delta \in (0,1)$.
\end{assumption}
While \Cref{ass:base-error-estimate} relies on the probabilistic bound $\hatu(S_{\err}, h, \delta)$ to hold for each $h \in \calH$, \Cref{ass:specific-instance-error-estimate} only requires a specific probabilistic bound for the instance $h^*$. However, since $h^*$ is defined in terms of unknown estimands $\theta_h$, we cannot directly compute $\hatu(S_{\err}, h^*, \delta)$. Instead, our strategy is to use \Cref{ass:h*bound} to construct a non-increasing sequence of classes $\{\calH_k(S_\err)\}_{k \in \mathbb{N}}$ which will all contain $h^*$ under certain conditions. We arrive at the following corollary, see \Cref{sec:local_proof} for detailed proof.

\begin{corollary}
    \label{cor:localize}
    Fix $h^* \in \calH$ and suppose Assumptions \ref{ass:h*bound} and \ref{ass:specific-instance-error-estimate} hold for some constant $c$. Further, let $\calH_0 := \calH$ and define $\calH_k$, $\hatxi_k$ inductively as in \eqref{eq:localizers_def}. 
    \begin{equation}
        \label{eq:localizers_def}
        \begin{aligned}
            \hatxi_k(S_\err, \delta) &:=  \max_{h \in \calH_k(S_\err)}\hatu(S_\err, h, \delta)\\ \calH_{k+1}(S_\err) &:= \left\{h \in \calH| -\estimate_{h} \le \hatxi_{k}(S_\err, \delta) -c\right\}
        \end{aligned}
    \end{equation}
    Then, for any $\delta \in (0,1)$, we have \eqref{eq:localized_result} holds.
    \begin{equation}
    \label{eq:localized_result}
        \Pr_{(S_{\est},S_{\err})}\Big( \hatxi_k(S_{\err} ,\delta)  \ge e_{h^*} \quad \forall k \in \mathbb{N}\Big)\geq 1-\delta
    \end{equation}
\end{corollary}

\section{Error Estimation Examples}
\label{sec:examples}
The goal of this section is to better understand error estimation. We outline a simplified recipe (\Cref{sec:ee_framework}), followed by two examples: bounding error for multiple mean estimates (\Cref{sec:ci-multiple-mean}), and bounding excess risk in supervised learning tasks (\Cref{sec:excess-risk-error-estimation}). The latter showcases how error estimation can replace complexity-based bounds, and how localization can further refine them.
\subsection{Simplified Recipe for Error Estimation}
\label{sec:ee_framework}
Let $S_\est$ and $S_\err$ be the error defining and error estimating datasets respectively. For simplicity of exposition, we focus on errors of the form $e_h = \theta_h - \estimate_h$, and detail how to construct explicit bounds $\hatu(S_{\err}, h, \delta)$ to satisfy \Cref{ass:base-error-estimate}. 
\begin{table}[h]
\centering
\small
\caption{Key quantities used for an estimand $\theta_h$.}
\renewcommand{\arraystretch}{1.2}
\begin{tabular}{c|cc}
\toprule
 & \textbf{Defining data} $S_{\text{est}}$ & \textbf{Error estimation data} $S_{\text{err}}$ \\ \midrule
Estimate              & $\displaystyle\hat\theta_{h}$          & $\displaystyle\hat\theta_{\text{err},h}$ \\
Error                 & $\displaystyle e_h=\theta_h-\hat\theta_{h}$ & $\displaystyle e_{\text{err},h}= \theta_h-\hat\theta_{\text{err},h}$ \\
\bottomrule
\end{tabular}
\label{tab:key_ee_terms}
\end{table}
\begin{enumerate}
    \item \textbf{Generate $S_{\err}$ estimates.} In addition to the tuple $(\theta_h,\estimate_{h},e_{h})$ from $S_{\est}$, we use $S_{\err}$ to define estimates $\estimate_{\err, h}$ and errors $e_{\err, h}$ (see \Cref{tab:key_ee_terms}). Now, for each $h \in \calH$,
\begin{equation}
    \label{eq:oneline_ee}
        e_{h} =  \theta_{h} - \estimate_{\err,h} + \estimate_{\err, h} - \estimate_{h} = e_{\err, h} + \estimate_{h} - \estimate_{\err,h}
\end{equation}

By construction, $e_h$ and $e_{\err,h}$ are independent, so an instance $h'$ determined by a realization of $S_\est$, can be treated as a single fixed instance in $\calH$ with respect to the randomness of $S_\err$.

\item \textbf{Construct $\hatu$ to satisfy \Cref{ass:base-error-estimate} (or \ref{ass:specific-instance-error-estimate})} by applying a single high-probability bound to $e_{\err, h}$. If $b_{\delta,h}$ is a $1-\delta$ probability upper bound on $e_{\err, h}$, we set \[\hatu(S_\err, h, \delta) := b_{\delta, h} + \estimate_{h} - \estimate_{\err,h},\]  Now, conditioning on $s$, under the $(1-\delta)$-probable event  $\{ e_{\err, h'} \le b_{\delta, h'} \}$, we have
\begin{equation}
    \label{eq:sf_error_bound}
        e_{h'} \leq b_{\delta, h'} + \estimate_{h'} - \estimate_{\err,h'} =: \hatu(S_\err, h', \delta)
\end{equation}

Thus we satisfy \Cref{ass:base-error-estimate}, and can set the max error bound $\max_{h \in \calH}\hatu(S_\err, h, \delta)$.
\item (Optional) \textbf{Localize the bound.} In the setting from \Cref{sec:localizing}, we can improve our error bound by \Cref{cor:localize}. We define $\calH_0 = \calH$, and $\calH_k$, $\hatxi_k$ as in \eqref{eq:localizers_def}. This gives
\begin{align}
\label{eq:sf_local_error_bound}
\Pr_{S\err \sim D}(\forall k \in \mathbb{N}, \hatxi_k(S_\err, \delta)  \ge e_{h^*} |\calE(s)) \ge 1-\delta
\end{align}
 In practice, the bounds $\hatxi_k$ converge as $k \to \infty$.
\end{enumerate}
We make a couple of remarks about our resulting method.

\begin{remark}[Relationship between $\hatu$ and e]
\label{rmk:general-surrogate}
     By construction: \[\hatu(S_{\err}, h, \delta) = b_{\delta, h} + \estimate_{h} - \estimate_{\err,h} = b_{\delta,h} + e_{\err, h} - e_h,\]
     Thus, when $e$ and $e_{\err}$ are i.i.d. random variables, $\hatu$ inherits a closely related distribution to $e$.\footnote{For example, if $e_h \sim \mathcal{N}(0,1)$, then $\hatu(S_{\err}, h, \delta) \sim \mathcal{N}(b_{\delta, h}, 2)$} Moreover, $\hatu$ preserves the dependence pattern across instances $h \in \calH$. In particular, if $b_{\delta,h}$ is constant across $h \in \calH$ (as in examples below), then $\text{Cov}(\hatu(S_{\err}, h, \delta) , \hatu(S_{\err}, h', \delta) ) = 2\text{Cov}(e_h, e_{h'})$. Thus $\hatu(S_{\err}, h, \delta)$ is a surrogate for $e_h$, so $\max_{h \in \calH}\hatu(S_{\err}, h, \delta)$ often closely tracks $\max_{h \in \calH}e_h$.
     \end{remark}

\begin{remark}[There is Room for Improvement]
\label{rmk:room-for-improvement}
In an ideal world, $\hatu(S_{\err},h,\delta)$ would be identically distributed as $e_{h}$ (while still being a valid upper bound). Our construction for $\hatu(S_{\err},h,\delta)$ often leads to twice as much variance as $e_{h}$ and we need twice as much data to define the same error. Hence, compared to the ``ideal" scenario, our error estimation intervals are larger by a factor of $\sqrt{4}=2$. Cross-fitting can substantially reduce this sample-splitting inflation while retaining validity; see Appendix~\ref{app:crossfit}.
\end{remark}

\begin{figure}
    \centering
    \includegraphics[width=\linewidth]{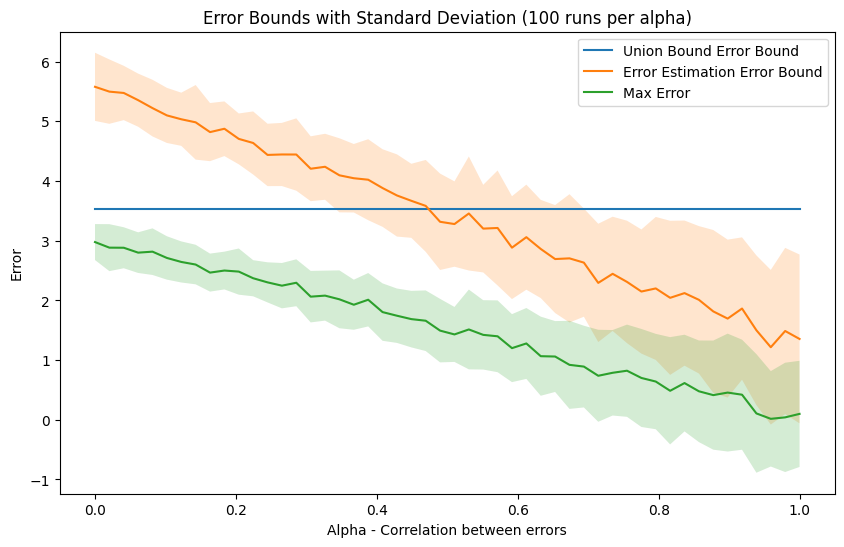}
    \caption{Error estimation vs. union bound. As correlation between tasks increases, the true max error (green) decreases. Our bound (orange) adapts accordingly while the union bound (blue) stays fixed due to its explicit dependence on $|\calH|$. See Appendix~\ref{app:exp_details} for details.}
    \label{fig:finite-sample}
\end{figure}

\begin{remark}[Tightness compared to union bound]
\label{rmk:ee-tightness}
     Since $\hatu(S_{\err}, h, \delta)$ is a set of random quantities, a union bound over $h \in \calH$ can be applied to (loosely) bound $\max_h \hatu(S_{\err}, h, \delta)$. Thus, in finite cases, our method is always no worse than the union bound up to constant factors. In practice, when errors are correlated, our method can provide significantly tighter bounds (see \Cref{fig:finite-sample}).
\end{remark}

\subsection{Confidence Interval Construction for Multiple Mean Estimates}
\label{sec:ci-multiple-mean}
Let $Y:\calH\rightarrow [-M,M]$ be a bounded random map (such as subgroups to average test score), and define $\theta_h:=\E_{Y\sim \calD}[Y(h)]$ for all $h\in\calH$. Our goal is to construct simultaneously valid CIs for these estimands that hold jointly with probability $1-\delta$.\footnote{See Appendix~\ref{app:ee_example_details} for a conversion of Example 1 to the setting in \Cref{sec:ci-multiple-mean}.} We draw $2n$ i.i.d. samples, and split them randomly into two independent datasets of equal size (for simplicity), $S_{\est}$ and $S_{\err}$. We define central quantities in \Cref{tab:multiple-mean}.

\begin{table}[ht]
\centering
\small
\caption{Error-Estimation Quantities for Multiple Means}
\begin{tabular}{llll}
\toprule
\textbf{Term} & \textbf{Definition} & \textbf{Term} & \textbf{Definition} \\
\midrule
$\estimate_{h}$ & $\sum_{y\in S_\est} \frac{y(h)}{n}$ & $\estimate_{\err,h}$ & $\sum_{y\in S_\err} \frac{y(h)}{n}$ \\
$\hat{\sigma}_{h}^2$ & $\sum_{y\in S_\est} \frac{(y(h)-\estimate_{h})^2}{n-1}$ & $\hat{\sigma}_{\err,h}^2$ & $\sum_{y\in S_\err} \frac{(y(h)-\estimate_{\err,h})^2}{n-1}$ \\
$e_{h}$ & $\frac{\sqrt{n}|\estimate_{h} - \theta_h|}{\hat{\sigma}_{h}}$ & $e_{\err,h}$ & $\frac{\sqrt{n}|\estimate_{\err,h} - \theta_h|}{\hat{\sigma}_{\err,h}}$ \\
\bottomrule
\end{tabular}
\label{tab:multiple-mean}
\end{table}
For any $h\in\calH$, $\estimate_{h}$ is our mean estimate around which we will construct our CI. The error $e_{h}$ is the absolute difference between this estimate and the target estimand ($|\theta_h-\estimate_{h}|$) divided by the sample standard error ($\hat{\sigma}_{h}/\sqrt{n}$).  By constructing a high-probability upper bound on the max error ($\max_{h\in\calH} e_{h}$), we get simultaneously valid high-probability CIs for all our estimands. We establish $\estimate_{\err,h}$, $\hat{\sigma}_{\err,h}$, and $e_{\err,h}$ from dataset $S_{\err}$ (see \Cref{tab:multiple-mean}).

In this section, to derive an estimate $\hatu$ satisfying \Cref{ass:base-error-estimate}, we implement a central limit theorem (CLT)-based asymptotic guarantee.\footnote{We use CLT due to its popularity, however, non-asymptotic alternatives, such as Hoeffding or Bernstein's inequality would lead to finite-sample guarantees.} By construction, we have defined $e_h$ and $e_{\err, h}$ to be distributed as $|\mathcal{N}(0,1)|$ random variables. Thus, from the CLT, $z_{\delta/2}$ upper bounds $e_{\err,h}$ asymptotically with probability at least $1-\delta$, where $z_{\delta}$ is the $\delta$ critical value for the standard normal distribution ($N(0,1)$). Now, if indeed $e_{\err,h} \leq z_{\delta/2}$ holds, then
\begin{equation}
\label{eq:asymptotic-hatu-for-multimean}
\begin{aligned}
    e_{h} &= \frac{\sqrt{n}|\estimate_{h} - \theta_h|}{\hat{\sigma}_{h}}\\
    &\leq \frac{\sqrt{n}|\estimate_{h} - \estimate_{\err,h}|}{\hat{\sigma}_{h}} + \frac{\hat{\sigma}_{\err,h} z_{\delta/2}}{\hat{\sigma}_{h}}=: \hatu(S_{\err},h,\delta) 
\end{aligned}
\end{equation}
 Hence, for all $h$, \[\Pr_{S_{\err}}(\hatu(S_{\err},h,\delta)\geq e_{h}|S_{\est})\geq 1-\delta\]
 asymptotically holds, so for $n$ is large enough for CLT-based CIs to hold, we have satisfied \Cref{ass:base-error-estimate}. From \Cref{thm:max-error-estimation-result}, we have the following corollary.
\begin{corollary}
\label{cor:ci-multiple-mean}
    Consider the setup in \Cref{sec:ci-multiple-mean}. Suppose the following CLT-based CIs hold: \[\Pr_{S_{\err}}(\hatu(S_{\err},h,\delta)\geq e_{h}|S_{\est})\geq 1-\delta\] for all $h\in\calH$. Then for $\hatxi(S_{\err},\delta):=\max_{h\in\calH} \hatu(S_{\err},h,\delta)$, we have \eqref{eq:multi-ci-bound-guarantee} holds.
    \begin{equation}
    \label{eq:multi-ci-bound-guarantee}
        \begin{aligned}
            \Pr_{\bar{S}}\left(\forall h\in\calH,\; |\estimate_{h} - \theta_h|\leq \frac{\hat{\sigma}_{h}}{\sqrt{n}}\hatxi(S_{\err},\delta)\right)\geq 1-\delta
        \end{aligned}
    \end{equation}
\end{corollary}

We are able to construct simultaneously valid CIs without requiring explicit union-bound-based adjustments. Using $\hatxi(S_{\err},\delta)$ in our CIs instead of $z_{\delta/2}$ is a data-driven adjustment for constructing multiple simultaneously valid CIs. This can be much better than the union-bound-based correction of $z_{\delta/(2|\calH|)}$ when $\calH$ is finite. To see the benefits of the above data-driven adjustment, see \Cref{fig:finite-sample}.

\subsection{Supervised Learning: Excess Risk Estimation and Constructing Set of Potentially Optimal Models}
\label{sec:excess-risk-error-estimation}

We consider supervised learning with data $(X, Y) \sim \calD$ over $\calX \times \calY$, a model class $\calG \subseteq \calX \to \calY$, a bounded loss $l: \calX \times \calY \times \calG \to [0, M]$, and two independent datasets $S_{\est}$ and $S_{\err}$ of $n$ i.i.d. samples each from $\calD$. The risk of a model $g$ is $\calR(g) := \E_{\calD}[l(X,Y,g)]$. Let $g^* = \arg\min_{g \in \calG} \calR(g)$ denote the best-in-class model; the excess risk $\Delta(g) := \calR(g) - \calR(g^*)$ quantifies how suboptimal $g$ is.

Given a current model $g_{\est}$, an upper bound on the excess risk ($\Delta(g_{\est})$) would help determine if we should collect more data for learning from the class $\calG$. A large bound implies room for improvement, while a small bound indicates the model is already close to optimal. Bounding excess risk typically requires complex theoretical tools. Using the localization technique introduced in \Cref{sec:localizing}, we construct excess risk bounds and thus define a set of potentially optimal models -- that is, a set which will contain $g^*$ with high-probability.\footnote{Constructing a set of potentially optimal models is the primary statistical step when constructing confidence intervals for UCB (upper confidence bound) based contextual bandit/RL algorithms.}

We formulate the problem of upper bounding the excess risk ($\Delta(g_{\est})$) as an error estimation problem. For $g\in\calG$, let $\theta_g:=\calR(g_{\est}) - \calR(g)$ be a measure of the sub-optimality of $g_{\est}$, when compared against a candidate model $g$. Therefore, the excess risk $\Delta(g_{\est})$ is equal to $\theta_{g^*}$.  We define estimates of $\theta_g$ from $S_\est$ and $S_\err$, as outlined in \Cref{tab:SL}.
\begin{table}[ht]
\centering
\caption{Error Estimation Quantities for Excess Risk Estimation}
\begin{tabular}{ll}
\toprule
\textbf{Term} & \textbf{Definition} \\
\midrule
$\theta_g$ & $\calR(g_{\est}) - \calR(g) = \E_D[l(X,Y,g_{\est})-l(X,Y,g)]$ \\
$\estimate_{g}$ & $(1/n)\sum_{(x,y)\in S_{\est}}(l(x,y,g_{\est})-l(x,y,g))$ \\
$\estimate_{g,\err}$ & $(1/n)\sum_{(x,y)\in S_{\err}}(l(x,y,g_{\est})-l(x,y,g))$ \\
$e_g$ & $\theta_g - \estimate_{g}$ \\
\bottomrule
\end{tabular}

\label{tab:SL}
\end{table}

Now we use Hoeffding's inequality, to construct a $\hatu$ which satisfies \Cref{ass:base-error-estimate}.
$$\hatu(S_{\err},g,\delta):=\estimate_{g,\err}-\estimate_{g}+2M\sqrt{\frac{\log(1/\delta)}{2n}}$$
 Defining $\hatxi_0(S_{\err},\delta):= \max_{g\in\calG}\hatu(S_{\err},g,\delta)$, by \Cref{thm:max-error-estimation-result}, \eqref{eq:sl_first_bound} holds with probability at least $1-\delta$.
\begin{equation}
    \label{eq:sl_first_bound}
    \begin{aligned}
        \Delta(g_{\est})= \theta_{g^*} = \estimate_{g^*} + e_{g^*} &\le \estimate_{g^*} + \hat{u}(S_{\err}, g^*, \delta)\\
        &\le \max_{g\in\calG} \estimate_{g} + \hatxi_0(S_{\err},\delta) 
    \end{aligned}
\end{equation}
By the definition of $g^*$, $\theta_{g^*} = \max_{g \in \calG}\theta_g$, and  $\theta_{g^*}(=\calR(g_{\est}) - \calR(g^*)) \ge 0$, so we have satisfied \Cref{ass:h*bound} with $c = 0$. We can thus apply \Cref{cor:localize} to improve the bound. Let $\calG_0 = \calG$, and define
\begin{equation}
\label{eq:supervised-learning-localizing}
    \begin{aligned}
        \hatxi_k(S_\err, \delta) &:= \max_{g \in \calG_k}\hatu(S_\err, g, \delta);\\
        \calG_k &:=\bigg\{g\in\calG_{k-1}\bigg| - \estimate_{g} \leq \hatxi_{k-1}(S_{\err},\delta) \bigg\};
    \end{aligned}
\end{equation}

For any $x\in\calX$, this set of potentially optimal models ($\calG_k$) allows us to construct a high-probability confidence set for $g^*(x)$, given by $\{g(x)|g\in \cap_{k \in \mathbb{N}} \calG_k\}$.\footnote{For some $K$, one would observe $\hatxi_{K+1} = \hatxi_{K}$, and stop the localization procedure.} We summarize our results in \Cref{cor:supervised-learning}. It is important to note that unlike popular theoretical bounds on excess risk \citep[e.g.,][]{koltchinskii2011oracle}, none of our results required an a priori analytic upper bound on class complexity of $\calG$ as an explicit input.

\begin{figure}[h]
    \centering
    \begin{minipage}[t]{\linewidth}
        \centering
        \includegraphics[width=0.9\linewidth]{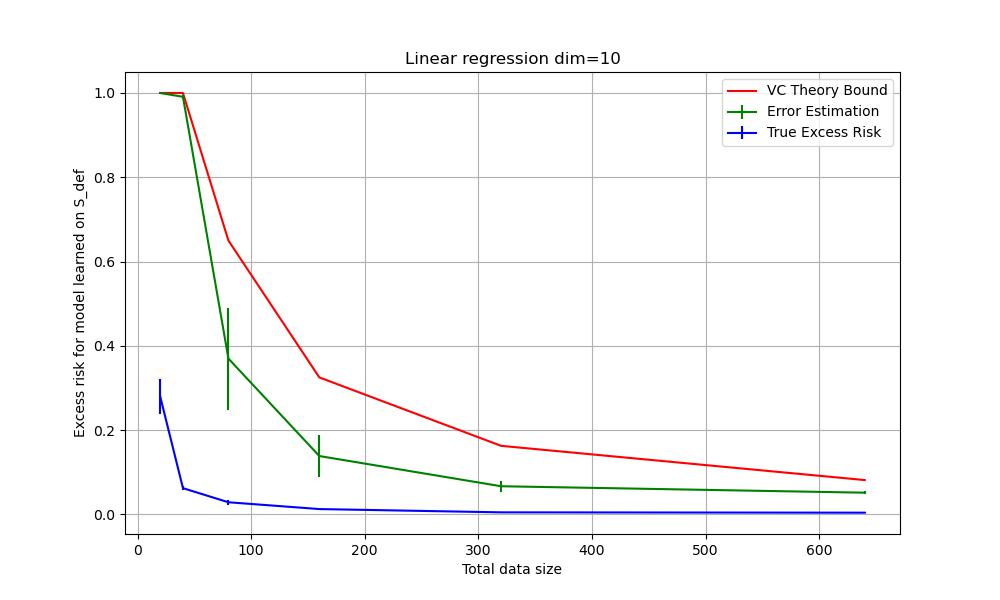}
        \caption*{(a) 10 dimensional linear regression model class, simulated datataset. Error estimated excess risk bounds (green) compared to VC theory excess risk bounds (red) in the linear case where VC theory is tight. True excess risk is shown in blue.}
        \label{fig:SL-lin}
    \end{minipage}
    
    \vspace{1em} 
    
    \begin{minipage}[t]{\linewidth}
        \centering
        \includegraphics[width=0.9\linewidth]{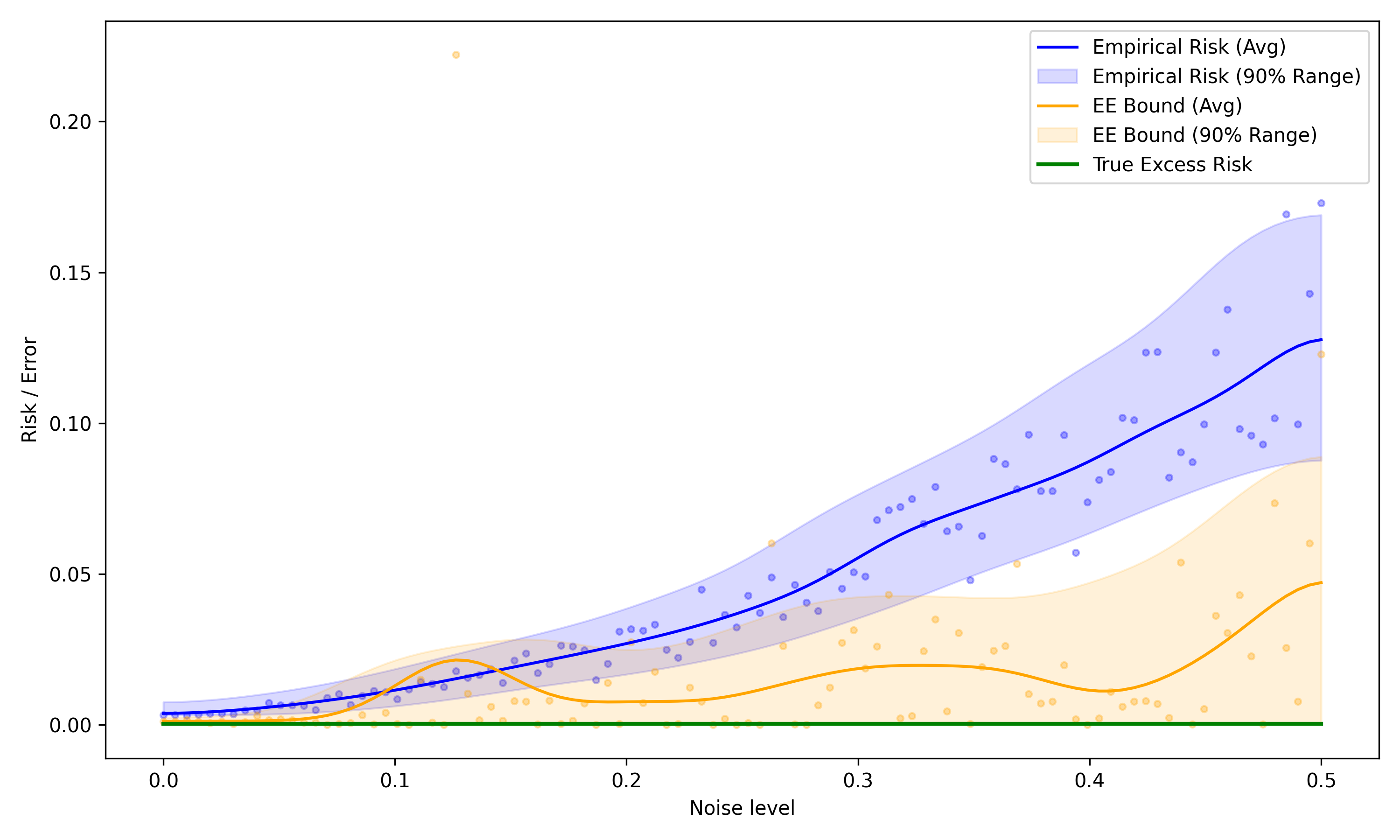}
        \caption*{(b) Neural networks with the structure [input dim - 128 - 64 - 32 - output dim] and ReLU activation. Error estimated excess risk bounds (orange) compared to the naive empirical risk (blue). True excess risk is shown in green.}
        \label{fig:SL-nn}
    \end{minipage}
    
    \caption{Performance of error-estimated excess risk bounds in different settings. See Appendix~\ref{app:exp_details} for simulation details. Figure 2(a) is based on synthetic data, Figure 2(b) uses the Iris dataset \cite{r_a_fisher_iris_1936}, downloaded from UCI.}
    \label{fig:SL-sim}
\end{figure}
\begin{corollary}
\label{cor:supervised-learning}
    Consider the supervised learning setting described in \Cref{sec:excess-risk-error-estimation}. With probability at least $1-\delta$, we have $\forall g\in\calG, e_g\leq \hatxi_0(S_{\err},\delta);$ and for all $k \in \mathbb{N}$, \eqref{eq:supervised-learning-guarantees} hold.
    \begin{equation}
    \label{eq:supervised-learning-guarantees}
        \begin{aligned}
            g^*\in \calG_k;\;\;\;&\Delta(g_{\est}) \leq \max_{g\in\calG_k} \estimate_{g} + \hatxi_k(S_{\err},\delta).
        \end{aligned}
    \end{equation}
\end{corollary}

\begin{remark}
    Note that a reasonable choice of $g_{\est}$ would be the minimizer $g_{\est} = \argmin_{g \in \calG}\mathbb{E}_{S_{\est}}[l(X,Y,g)]$. Then $\estimate_{g} \le 0$ and we could reduce our excess risk bound to $\Delta(g_{\est}) \leq  \min_k \hatxi_k(S_{\err},\delta)$.
\end{remark}
\Cref{fig:SL-sim} highlights two regimes:
(a) for linear models our bound coincides with the tight VC-theory curve, confirming no loss versus theory;
(b) for neural nets, as noise is added, our method still hugs the true excess risk (albeit with increased variance) while the naïve empirical-risk gap blows up. This demonstrates our method's ability to distinguish between excess risk and irreducible noise.

\section{Applications to Contextual Bandits}
\label{sec:apply_to_cbs}
In this section, we focus on the stochastic contextual bandit setting, further developed in Appendix~\ref{sec:cbs}. Contextual bandit algorithms require uncertainty estimates to effectively balance exploration and exploitation. In each round $t=1,\dots,T$, the learner observes a context $x_t\in\calX$, chooses an action $a_t\in\calA$, and observes a reward $r_t(a_t)$. The goal is to compete with the best policy in a class $\Pi\subseteq(\calX\to\calA)$:
\[
\pi^* \in \argmax_{\pi\in\Pi}\E_{(x,r)}[r(\pi(x))].
\]

Computationally tractable and theoretically sound bandit algorithms typically obtain exploration bonuses from error bounds on a reward/value model class $\calF$ (e.g., $f(x,a)\approx \E[r(a)\mid x]$). These bounds typically require a complexity measure of $\calF$ (VC/Rademacher/covering numbers) as an explicit input to the algorithm \citep{foster2023foundations,agarwal2014taming,foster2020instance}. For many model classes $\calF$ (e.g., deep nets, tree-based policies), theoretically deriving complexity bounds is hard and often conservative. 

We show error estimation provides a drop-in, data-driven substitute: it yields the needed confidence widths without requiring the complexity of $\calF$ as input. We first plug it into a simple inverse gap weighting (IGW) method in the linear setting (where tight theory exists under realizability assumptions\footnote{i.e. the assumption that the true reward model $f \in \calF$.}). We then (\Cref{subsec:cb_pipeline}) outline a theoretically sound and modular pipeline where reward models can be estimated via any blackbox oracle\footnote{This oracle may not be related to the policy class of interest, for example the oracle could select the best reward model from multiple classes via cross validation.} and error estimation is used for constructing two forms of uncertainty sets (conformal arm sets and arm elimination sets) to quantify the error of the reward model in terms of evaluating the policy class of interest without assuming realizability. Details and guarantees for this pipeline are provided in Appendix~\ref{sec:cbs}.

\subsection{Improving FALCON (IGW) via error estimation}
\label{sec:falcon}

\begin{figure}[t]
    \centering
    \includegraphics[width=\linewidth]{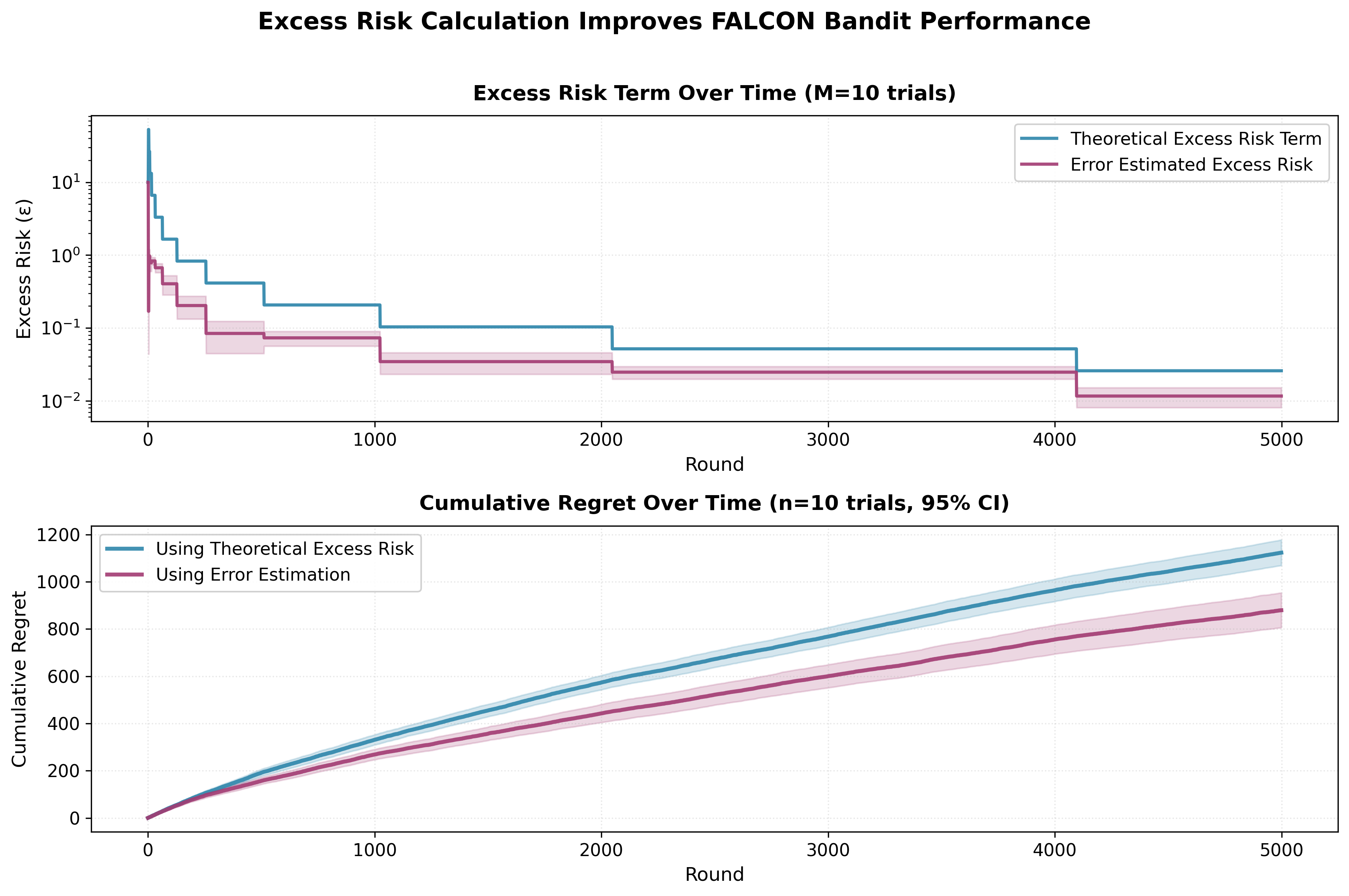}
    \caption{Linear contextual bandit: FALCON with standard linear uncertainty vs.\ FALCON with error-estimated uncertainty. The top plot shows excess risk bounds using theoretical bounds vs. error estimation. The bottom plot shows performance of algorithms based on said excess risk bounds. See Appendix~\ref{app:exp_details} for details.}
    \label{fig:falcon_linear_regret}
\end{figure}

Many contextual bandit algorithms follow the same template: fit a reward model from past data, form uncertainty estimates, and construct a policy that balances exploration vs exploitation. A representative example is FALCON \citep{simchi2020bypassing}, which fits a regressor $\hat f_t$ for $\E[r\mid x,a]$ and quantifies uncertainty via a theoretical excess risk bound over the underlying reward model class -- as the excess risk bound shrinks, FALCON samples arms with higher estimated reward more often. 

By using data from earlier epochs as $S_{\est}$ and data from the last epoch as $S_{\err}$, we compute a high-probability upper bound $\hat\xi$ on the excess risk of the current estimated reward model. As this bound is instance-dependent, it tends to be tighter, thus using it in FALCON results in lower cumulative regret (Figure~\ref{fig:falcon_linear_regret}). We illustrate this in a linear setting, where standard theory is already well understood; we expect the benefit to be larger for richer model classes where complexity bounds are often looser.

\subsection{A Modular Contextual Bandit Pipeline}
\label{subsec:cb_pipeline}

Figure~\ref{fig:cb_pipeline_teaser} summarizes the main idea behind our contextual bandit pipeline (full details and guarantees in Appendix~\ref{sec:cbs}). We treat reward modeling and policy evaluation as black-box oracles; error estimation is the shared uncertainty framework, which we use in two complementary ways: (i) to build an elimination set $\hat{g}(x)$ of actions that contain $\pi^*(x)$ with high probability\footnote{$\hat{g}(x)$ is similar in goal and guarantee of constructing candidate action sets in \cite{foster2020instance}, although we do not assume realizability and our approach may be of independent interest.}, and (ii) to construct conformal arm sets $C(x,\zeta)$ which contain $\pi^*(x)$ for at least $1-\zeta$ fraction of contexts with high probability \citep{krishnamurthy2023proportional}. These sets can then be combined with different exploration rules (e.g., see \eqref{eq:define_pma_genRAPR} in \Cref{alg:GenRAPR}), enabling a plug-and-play design where improving any component does not require re-deriving guarantees for the overall pipeline.

In addition to the modularity and not requiring mathematically derived complexity bounds as an algorithmic input, a key benefit is that the pipeline does not require realizability: we do not assume that the true conditional expected reward model ($\E[r(a)\mid x]$) belongs to our reward estimation class $\calF$. Instead, error estimation is used to evaluate how well a given reward model estimates differences in policies within the class of interest\footnote{Similar to the measure of error used in misspecification tests by \cite{krishnamurthy2024towards}.} \footnote{Further, since we only evaluate if our reward model can differentiate between policies in the class of interest, our reward models can be estimated via heterogeneous treatment effect estimators and enjoy the corresponding statistical benefits (see, e.g. \cite{carranza2023flexible}).}.

\begin{figure}[t]
\centering

\begin{tikzpicture}[
  font=\scriptsize,
  node distance=2.6mm,
  >={Stealth[length=2mm]},
  box/.style={draw, rounded corners=2pt, thick, inner sep=2.5pt, minimum height=4.2mm},
  arr/.style={->, thick, shorten >=1pt, shorten <=1pt},
  txt/.style={align=center},
  data/.style={box, fill=blue!12, draw=gray!55},
  oracle/.style={box, fill=gray!20, draw=black!55},
  ee/.style={box, fill=orange!30, draw=orange!55},
  set/.style={box, fill=orange!30, draw=black!55},
  output/.style={box, fill=blue!12, draw=gray!55},
  note/.style={txt, text=gray!70}
]
  \node[data, txt, text width=0.78\columnwidth] (S) {Data from past epochs $S$};

  \newlength{\twoboxgap}
  \setlength{\twoboxgap}{4mm}
  \newlength{\totalw}
  \setlength{\totalw}{0.78\columnwidth}
  \newlength{\halfboxw}
  \setlength{\halfboxw}{0.5\dimexpr \totalw-\twoboxgap\relax}
  \newlength{\pairdx}
  \setlength{\pairdx}{0.5\dimexpr \halfboxw+\twoboxgap\relax}

  \node[inner sep=0pt, minimum height=4.2mm, below=of S] (splitc) {};
  \node[data, txt, text width=\halfboxw] (Sdef) at ([xshift=-\pairdx]splitc) {$S_{\est}$};
  \node[data, txt, text width=\halfboxw] (Serr) at ([xshift= \pairdx]splitc) {$S_{\err}$};

  \node[oracle, txt, text width=0.75\halfboxw, below=of Sdef] (oracles)
    {black-box oracles\\[-1pt]
     \textcolor{black!60}{$\hat R,\hat\pi, \hat f$}};

  \node[ee, txt, text width=0.70\totalw, below=6mm of oracles, xshift=\pairdx] (ee)
    {error estimation\\[-1pt]\textcolor{red!70}{$\hat U_{\text{arm elimination}}$ \;\; and \;\; $\hat U_{\text{conformal arm sets}}$}};

  \node[set, txt, text width=\totalw, below=of ee] (sets)
    {sets for exploration\\[-1pt]
     \textcolor{black!60}{elimination $\hat g(x)$} \;\; + \;\;
     \textcolor{black!60}{conformal $C(x)$}};

  \node[output, txt, text width=\totalw, below=of sets] (p)
    {exploration policy $p_{\tau+1}$};

  \draw[arr] (S) -- (Sdef);
  \draw[arr] (S) -- (Serr);

  \draw[arr] (Sdef) -- (oracles);

  \draw[arr] ([xshift=-3mm]Sdef.south east) -- (ee);
  \draw[arr] (Serr) -- (ee);

  \draw[arr] (oracles.south) -- (ee);
  \draw[arr] ([xshift=4mm]oracles.south west) -- ([xshift=8mm]sets.north west);


  \draw[arr] (ee) -- (sets);
  \draw[arr] (sets) -- (p);

\end{tikzpicture}

\caption{Bandit pipeline: error estimation calibrates uncertainty used for both elimination and conformal arm sets, allowing for a modular approach with no realizability assumption required. See Appendix~\ref{sec:cbs} for algorithm details.}
\label{fig:cb_pipeline_teaser} 
\end{figure}
\section{Conclusion}

The problem of bounding the maximum error for a class of estimand-estimate-error tuples (i.e. error estimation problem) is fundamental to statistical learning theory and is applicable to a wide range of important problems. In this paper we circumvent previous challenges of solving the error estimation problem by introducing a single optimization procedure, without needing to mathematically analyze the complexity of the underlying class. Our procedure can further be combined with localization arguments to achieve tighter error bounds for specific hypotheses of interest (e.g. in supervised learning, we are particularly interested in the hypothesis that minimizes the true risk). Finally, we demonstrate the value of this technique through various applications (see Sections \ref{sec:examples} and \ref{sec:apply_to_cbs}).

\textbf{Limitations:} While our method avoids the need for theoretical complexity bounds, there are practical considerations. Firstly, our approach relies on an explicit data split, which inflates intervals by reserving half the samples for error estimation. Cross-fitting or other k-fold schemes might reclaim those samples (see Appendix \ref{app:crossfit}) and merits further study. Secondly, we tackle solving $\sup_{h \in \calH}\hatu(S_\err, h, \delta)$ via SGD methods (Adam). However, it may be better tackled via DC programming solutions\footnote{Note $-\hatu$ is often a difference of convex functions} \cite{horst_dc_1999}.

\bibliography{ref}
\bibliographystyle{icml2026}

\onecolumn
\include{icml2026/appendix}
\include{icml2026/appendix_bandits_pipeline}


\end{document}

%% file: icml2026/appendix.tex

\newpage
\appendix
\counterwithin{theorem}{section}

\renewcommand{\thetheorem}{\thesection.\arabic{theorem}}

The appendix is structured as follows. 
\begin{itemize}
    \item \Cref{app:rad} formalizes the connection to Rademacher complexity, relating data splitting to symmetrization. Moreover, we prove tightness of Error Estimation to Rademacher complexity and discuss its relative advantages.
    \item \Cref{app:overfitting} offers some intuition for error estimation by outlining an analogy to the classical overfitting inequality in supervised learning.
    \item \Cref{app:exp_details} contains details for how each of the plots was generated as well as information about the compute infrastructure.
    \item \Cref{sec:local_proof} details how to extend the main theorem from maxima to suprema in the continuous case. We also prove \Cref{cor:localize}.
    \item \Cref{app:ee_example_details} provides additional details for \Cref{sec:examples}. We tie Example 1 from the introduction to the setting in \Cref{sec:ci-multiple-mean}, and we prove that in this setting, the distribution of our constructed $\hatu(S_\err, h, \delta)$ is a scaled version of the distribution of $e_h$. 
    \item \Cref{sec:multiple-hyps} details an application of error estimation to multiple hypothesis testing.
    \item \Cref{app:crossfit} explores how error estimation can be further improved using crossfitting techniques
    \item \Cref{sec:cbs} outlines an implementation of contextual bandit algorithms relying on the error estimation theory presented in this paper.
\end{itemize}

\section{Connection to Rademacher Complexity}
\label{app:rad}
\subsection{Data Splitting as Rademacher Symmetrization}
Traditionally, excess risk is bounded via generalization error, which in turn is bounded using Rademacher complexity. We recall the result from \cite{goos_rademacher_2001} below.
\begin{theorem}
        Let $X_{1},\dots,X_{n}$ be \emph{i.i.d.}\ draws from a distribution $P$ on some space $\mathcal{X}$, and let $\mathcal{F}$ be a class of bounded functions $[-b,b]$. Then, for any $\delta \ge 0$, with probability at least $1 - \exp(-\delta^2 n / 2b^2)$,
        \[
  \bigl\|P_{n}-P\bigr\|_{\mathcal{F}} =  \sup_{f\in\mathcal{F}}  \Bigl|\tfrac1n\sum_{k=1}^{n}f(X_{k}) - \mathbb{E}[f(X)]\Bigr| \le  2\mathcal{R}_{n}(\mathcal{F}) +\delta.
\]  
    \end{theorem}

\begin{proof}
The argument breaks into two parts:

\medskip
\noindent\textbf{(1) Concentration via McDiarmid’s inequality.}\\
Since each $f(X_k)$ lies in an interval of length $2b$, the functional 
\[
  S(X_{1:n}) \;=\; \bigl\|P_{n}-P\bigr\|_{\mathcal{F}}
\]
changes by at most $2b/n$ when any single $X_k$ is replaced.  McDiarmid’s bounded-differences inequality thus gives
\[
  \Pr\bigl\{S - \mathbb{E}S \ge \delta\bigr\}
  \;\le\;\exp\!\Bigl(-\tfrac{2n\,\delta^{2}}{(2b)^{2}}\Bigr)
  =\exp\!\Bigl(-\tfrac{\delta^{2}n}{2b^{2}}\Bigr).
\]
Hence with probability $\ge1-\exp(-\frac{\delta^2n}{2b^2})$,
\[
  \bigl\|P_{n}-P\bigr\|_{\mathcal{F}}
  \;\le\;
  \mathbb{E}\bigl\|P_{n}-P\bigr\|_{\mathcal{F}}
  \;+\;\delta.
\]

\medskip
\noindent\textbf{(2) Bounding the expectation by Rademacher complexity.}\\
We introduce \emph{ghost} samples $Y_{1},\dots,Y_{n}\overset{\mathrm{iid}}{\sim}P$, independent and identically distributed to the $X_k$. Then,
\begin{align}
  \mathbb{E}\bigl\|P_{n}-P\bigr\|_{\mathcal{F}}
  &=\;\mathbb{E}_{X}\Bigl[\sup_{f\in\mathcal{F}}
     \bigl|\tfrac1n\sum_{k}f(X_{k}) - \mathbb{E}f(Y)\bigr|\Bigr] \nonumber\\
  &\stackrel{(i)}{\le}\;\mathbb{E}_{X,Y}\Bigl[\sup_{f\in\mathcal{F}}
     \bigl|\tfrac1n\sum_{k}\bigl(f(X_{k}) - f(Y_{k})\bigr)\bigr|\Bigr] \label{eq:ghost_expectation}\\
  &\stackrel{(ii)}{\le}\;\mathbb{E}_{X,Y,\epsilon}
     \Bigl[\sup_{f\in\mathcal{F}}\bigl|\tfrac1n\sum_{k}
     \epsilon_{k}\,(f(X_{k})-f(Y_{k}))\bigr|\Bigr] \nonumber\\
  &\stackrel{(iii)}{\le}\;2\;\mathbb{E}_{X,\epsilon}
     \Bigl[\sup_{f\in\mathcal{F}}\bigl|\tfrac1n\sum_{k}
     \epsilon_{k}\,f(X_{k})\bigr|\Bigr]
     \;=\;2\,\mathcal{R}_{n}(\mathcal{F}). \nonumber
\end{align}
\begin{itemize}
  \item[(i)] uses the i.i.d.\ nature of $Y_k$ to replace $\E[f(Y)]$ by the average on the ghost samples, and Jensen's inequality to bring the expectation outside of the absolute value and the supremum.
  \item[(ii)] is a symmetrization via Rademacher signs $\epsilon_k\in\{\pm1\}$.
  \item[(iii)] applies the triangle inequality $|f(X_k)-f(Y_k)|\le|f(X_k)|+|f(Y_k)|$ and the fact that both $X$ and $Y$ are identically distributed.
\end{itemize}
Combining (1) and (2) yields the stated bound.
\end{proof}

Our method of data splitting is analogous to collecting realizations of the ghost samples introduced in step (2). In fact, we use a very similar argument to show that our error estimation bound is bounded (with high probability up to constants) by Rademacher complexity. In the same style as argument (1), we apply McDiarmid's inequality to bound the error estimation bound directly by the expression in \Cref{eq:ghost_expectation}.
\subsection{Tightness of Error Estimation to Rademacher Complexity}
Consider the generalization bound setting in the Error Estimation framework. Draw two independent samples
\(
 S_{\est} = \{X_i\}_{i=1}^n
\)
and
\(
 S_{\err} = \{Y_i\}_{i=1}^n
\)
from $P$. We let $\calF$ be our class of functions, and define our estimands $P_f$, estimates $P_n f$ from $S_\est$ and alternative estimates $P_n^{\err}$ from $S_\err$.
\[ Pf = \mathbb{E}[f(X)],\quad
  P_nf = \frac1n\sum_{i=1}^n f(X_i),\quad
  P_n^{\err}f = \frac1n\sum_{i=1}^n f(Y_i).
\]
Our errors are defined as $|P_nf - Pf|$. By error-estimation (using Hoeffding's inequality), we have that with probability at least $1 - 2\exp\bigl(-\tfrac{\delta^2 n}{2b^2}\bigr)$, 
\begin{equation}
\label{eq:EEforGB}
    ||P_n - P||_\mathcal{F} := \sup_{f \in \mathcal{F}}\left|\frac1n\sum_{i=1}^n f(X_i) - \mathbb{E}[f(X)]\right|
\le \sup_{f \in \mathcal{F}}\left|\frac1n\sum_{i=1}^n f(X_i) - \frac1n\sum_{i=1}^n f(Y_i)\right| + \delta 
\end{equation}
Now we mimic the idea in (1), again using McDiarmid's inequality. This time, we define the functional \[S(X_{1:n}, Y_{1:n}) = \sup_{f \in \mathcal{F}}\left|\frac1n\sum_{i=1}^n f(X_i) - \frac1n\sum_{i=1}^n f(Y_i)\right|.\] Since we have $2n$ points but still the same bounded difference, we have with probability at least $1-\exp(-\frac{\delta^2n}{4b^2})$
\begin{equation}
\label{eq:McDforEE}
    \sup_{f \in \mathcal{F}}\left|\frac1n\sum_{i=1}^n f(X_i) - \frac1n\sum_{i=1}^n f(Y_i)\right| \le \mathbb{E}_{X,Y}\left[\sup_{f \in \mathcal{F}}\left|\frac1n\sum_{i=1}^n f(X_i) - \frac1n\sum_{i=1}^n f(Y_i)\right|\right] + \delta
\end{equation}
where the last two inequalities hold with high probability. The first by our error estimation procedure, and the second by McDiarmid's inequality. Note the term on the RHS is bounded directly by $2\mathcal{R}_n(\calF)$ as it appears in \Cref{eq:ghost_expectation}. Combining this with \eqref{eq:EEforGB} and \eqref{eq:McDforEE}, we have with probability $1 - \exp\bigl(-\tfrac{\delta^2 n}{4b^2}\bigr)$
\[\texttt{EE-bound}:=\sup_{f \in \mathcal{F}}\left|\frac1n\sum_{i=1}^n f(X_i) - \frac1n\sum_{i=1}^n f(Y_i)\right| + \delta \le 2\mathcal{R}_n(\calF) + 2\delta\]

\subsection{Computational and distributional advantages}
Although in the generalization bound setting described above, our solution is analogous to introducing real "ghost" variables, our framework allows for weaker distributional assumptions. Recall that we do not require $S_\est$ and $S_\err$ to be identically distributed, thus allowing our solution to be especially suitable for the contextual bandit application, as outlined in \Cref{sec:cbs}.\\
Moreover, we escape the last step of the proof above and therefore we circumvent the expectation over all $\epsilon$ values as is introduced in the definition of Rademacher complexity. This is a key computational challenge that we significantly reduce.

\section{Overfitting analogy}
\label{app:overfitting}
To provide intuition for our approach, we draw an analogy between our solution to bounding maximum error over a class and the well-known overfitting phenomenon that often occurs in machine learning. The goal in many learning problems is to minimize true risk (expected loss over a data distribution) over a class of predictors $\calF$. Empirical risk minimization (ERM) typically approximates this by selecting $f \in \calF$ to minimize empirical risk (average loss) on some dataset $S$. This inherently ``overfits" to $S$: with high probability, the minimum empirical risk is less than the minimum true risk (up to a small correction).\footnote{This is closely related to the fact that training error is nearly always less than test error.} Formally, if we let $\calR(f)$ be the true risk and $\hat{\calR}(f)$ be the empirical risk, we see with high probability
\begin{equation}
 \min_{f \in \mathcal{F}} \hat{\mathcal{R}}(f) \lesssim \min_{f \in \mathcal{F}} \calR(f)
    \label{eq:overfitting}
\end{equation}
While overfitting is often viewed as an issue to be mitigated, we view it as an informative data-driven technique that can be used to construct a high probability lower bound on the minimum true risk (without ever needing to rely on class complexity bounds of $\calF$). 

Error estimation, like minimizing risk, involves optimizing an unknown quantity over a class. However, unlike fixed risks, errors are data-dependent random variables. To tackle this, we introduce a holdout dataset, that is not used for constructing our estimates and defining our errors. We then condition on the input estimates and error definitions, and are thus able to treat them as fixed quantities under the sampling distribution of the holdout dataset. Next we use the holdout dataset to construct high probability upper bounds on individual errors. \Cref{thm:max-error-estimation-result}, which is analogous to the overfitting inequality \eqref{eq:overfitting}, argues that the maximum of these high-probability upper bounds on individual errors is, in fact, a high-probability upper bound on the maximum error.
Thus, we can tackle error estimation without ever needing any knowledge of or bounds on the size/complexity of the input class.\\

\section{Details for Figures}
\label{app:exp_details}
\paragraph{Compute Infrastructure.}
All experiments were run locally on a MacBook Pro equipped with an Apple M3 Max chip (14-core CPU: 10 performance + 4 efficiency cores, 14 threads total; 30-core integrated GPU) and 36 GB of unified LPDDR memory. We recorded wall-clock and CPU-time using the Unix `/usr/bin/time` utility (flags “\%U user CPU seconds, \%S system CPU seconds, \%E elapsed real time”), and sampled GPU utilization via `powermetrics` (“GPU\%Busy”).

Per-experiment runtimes (wall-clock / CPU time / GPU-time) were:

\begin{itemize}
    \item Simulation for \Cref{fig:finite-sample}:  
    0.0005 h / 0.0004 h CPU / 0.0005 h GPU  
    \item Simulation for \Cref{fig:SL-sim} (a): 
    0.030 h / 0.029 h CPU / 0.030 h GPU 
  \item Simulation for \Cref{fig:SL-sim} (b):  
    0.63 h / 8.1 h CPU / 0.63 h GPU 
    \item Simultation for \Cref{fig:falcon_linear_regret}: 0.03h / 0.16h CPU / 0.03h GPU
\end{itemize}


\paragraph{\Cref{fig:finite-sample}}is generated as follows:
    \begin{itemize}
  \item \textbf{Point generation:} For each of 50 evenly spaced correlation levels $\alpha\in[0,1]$, draw two independent sets of $n=500$ points from $\mathcal{N}(0,1)$ with mixed correlation controlled by $\alpha$ (0 = independent, 1 = fully correlated).
  \item \textbf{Per-iteration procedure} (100 runs per $\alpha$):
    \begin{enumerate}
      \item Generate \texttt{points} and \texttt{err\_points} with the same $\alpha$.
      \item Compute the penalty term: 
        $\displaystyle \max_i\bigl(\texttt{points}_i - \texttt{err\_points}_i\bigr)\,$.
      \item Error-estimation bound:
        $\displaystyle \Phi^{-1}(1-\delta)
          + \max_i(\texttt{points}_i - \texttt{err\_points}_i)
          \quad\text{with }\delta=0.1\,$.
      \item Union bound:
        $\displaystyle \Phi^{-1}\bigl(1-\tfrac{\delta}{n}\bigr)\,$.
      \item Record one-sided max error:
        $\displaystyle \max_i(\texttt{points}_i)\,$.
    \end{enumerate}
  \item \textbf{Aggregation:} Compute the mean and $\pm1$\,SD of each bound and max error over the 100 runs at each $\alpha$.
  \item \textbf{Data-splitting note:} Using separate draws for penalty vs. error yields out-of-sample bounds but halves effective sample size, increasing variance.
  \item \textbf{One-sided coverage:} All metrics track only upper-tail excursions. For two-sided guarantees, replace maxima by $\max_i|\cdot|$ and use quantiles $\Phi^{-1}\bigl(1-\tfrac{\delta}{2n}\bigr)$.
\end{itemize}

\paragraph{\Cref{fig:SL-sim} (a)}is generated as follows: 
\begin{enumerate}
  \item \textbf{Data Generating Process (DGP):}
    \begin{itemize}
      \item Choose a 10-dimensional linear model by sampling coefficients $\beta$ i.i.d.\ from $\mathcal{N}(0,1)$ and scaling them to have $\|\beta\|_{2}=0.5$.
      \item For each data point, sample features $x\sim\mathcal{N}(0,I_{10})$ and normalize to $\|x\|_{2}=1$.
      \item Compute noiseless outcomes $y = x^\top \beta$ and add uniform noise $\varepsilon\sim \mathrm{Unif}[-0.5,0.5]$, so $y\in[-1,1]$ (enabling use of VC bounds requiring bounded labels).
    \end{itemize}

  \item \textbf{Training and Error Sets:}
    \begin{itemize}
      \item Fix one realization of the above DGP.
      \item For each dataset size $N$, sample $N$ points and split evenly into a \emph{def} set and an \emph{err} set.
      \item Train a linear predictor on the \emph{def} set; aim to bound its excess risk on new data.
    \end{itemize}

  \item \textbf{Evaluation and Plotting:}
    \begin{itemize}
      \item Compute:
        \begin{itemize}
          \item True excess risk,
          \item 95\% VC bound: $2\bigl(\tfrac{d + \ln(1/0.05)}{n_{\text{def}}}\bigr)$,
          \item 95\% error-estimation bound (Corollary 3.5).
        \end{itemize}
      \item Repeat over 5 independent simulations and average results.
      \item Plot each metric versus training size; observe that error estimation matches or slightly outperforms the VC bound in this linear setting.
      \item Note: for more complex hypothesis classes, error estimation is expected to yield larger improvements over classical VC bounds.
    \end{itemize}
\end{enumerate}

\paragraph{\Cref{fig:SL-sim} (b)}is generated as follows:
.\begin{enumerate}
  \item \textbf{Data Generating Process (DGP):}
    \begin{itemize}
      \item Use the UCI Iris dataset from \cite{r_a_fisher_iris_1936}: with features $X\in\mathbb{R}^{150\times 4}$ and one-hot encoded labels $Y \in \mathbb{R}^{150 \times 3}$.
      \item Train a ''true'' neural network $g^*$ on $(X,Y)$ with architecture  
        \[
          \text{input\_dim} \;\to\; 128 \xrightarrow{\text{ReLU}}
          64 \xrightarrow{\text{ReLU}}
          32 \xrightarrow{\text{ReLU}}
          \text{output\_dim} \xrightarrow{\text{Sigmoid}}
        \]
      \item Redefine labels via $Y_g = g^*(X)$ to make $g^*$ the "true" best in class (and for realizability).
      \item Add uniform noise $\varepsilon\sim\mathrm{Unif}[-\eta,\eta]$ to produce $Y'_g=Y_g+\varepsilon$.
    \end{itemize}

  \item \textbf{Training \& Splitting:}
    \begin{itemize}
      \item Fix one realization of $(X,Y'_g)$.
      \item Split into two equal halves:  
        \begin{itemize}
          \item $S_{\est}=(X_{\est},Y_{\est})$ for training the estimator $g_{\est}\equiv f$ (trained for 4000 epochs, batch size 8).
          \item $S_{\err}=(X_{\err},Y_{\err})$ for error evaluation.
        \end{itemize}
        \item  We train on $S_\est$ once to find $f$ and our goal is now to bound the excess risk of $f$.
        \item Compute true excess risk once as  
        \[
          R(f)-R(g^*) \;=\;\hat R(X,Y_g;f)\;-\;\hat R(X,Y_g;g^*)\,.
        \]
    \end{itemize}

  \item \textbf{Risk Evaluation:}
    \begin{itemize}
      \item Generate a grid of noise levels $\eta_k\in[0,0.5]$ (100 points).
      \item For each $\eta_k$, create noisy labels on both $S_{\est}$ and $S_{\err}$.
      \item Compute empirical risk $\hat{R}_{\est}^{\eta_k}(f)$ and $\hat{R}_{\err}^{\eta_k}(f)$ on each noisy $S_\est, S_{\err}$.
    \end{itemize}

  \item \textbf{Excess‐Risk Bound via Error Estimation (Cor.~3.5):}
    \begin{itemize}
      \item Instantiate \texttt{ExcessRiskClassic} with $(S_{\est},S_{\err},f,g^*)$, loss\,=\,$\ell_{MSE}$, and per‐sample loss vector.
      \item For each $\eta_k$, run the internal variational search (10 restarts) to minimize  
        \[
          \Delta_g := \hat{R}_{\err}^{\eta_k}(g)-\hat{R}_{\est}^{\eta_k}(g)\,+\,\text{(Hoeffding or conformal correction)}
        \]
        subject to $\hat{R}_{\est}^{\eta_k}(g)\le \hat{R}_{\est}^{\eta_k}(f)+\xi$.
      \item Localize $\xi$ until it stops decreasing, record bound $\xi$.
    \end{itemize}

  \item \textbf{Aggregation \& Visualization:}
    \begin{itemize}
      \item Repeat the excess‐risk bound computation across noise levels.
      \item Save results to \texttt{noise\_results\_updated.csv}.
      \item Plot for each $\eta_k$:  
        \begin{itemize}
          \item Scatter and 90\% range of $\hat R_{\err}(f)$,
          \item Scatter and 90\% range of $\xi$,
          \item True excess risk level.
        \end{itemize}
    \end{itemize}
\end{enumerate}

\paragraph{\Cref{fig:falcon_linear_regret}} is generated as follows:
\begin{enumerate}
  \item \textbf{Data Generating Process (DGP):}
    \begin{itemize}
      \item For each trial, sample true reward parameters $\theta_a \in \mathbb{R}^{10}$ independently for each action $a \in \{0,1,2,3,4\}$:
        \begin{itemize}
          \item Sample $\theta_a \sim \mathcal{N}(0, I_{10})$ and normalize to $\|\theta_a\|_2 \in [0.5, 2.0]$ (random scale).
        \end{itemize}
      \item For each round $t \in \{1, \ldots, T\}$ with $T=5000$:
        \begin{itemize}
          \item Sample context $x_t \sim \mathcal{N}(0, I_{10})$ and normalize to $\|x_t\|_2 = 1$.
          \item Compute noiseless reward $r_t(a) = x_t^\top \theta_a$ for each action $a$.
          \item Add Gaussian noise $\varepsilon_t \sim \mathcal{N}(0, 0.01)$ to observed rewards.
        \end{itemize}
      \item The same sequence of contexts and reward noise is used for both FALCON variants within each trial to ensure fair comparison.
    \end{itemize}

  \item \textbf{FALCON Algorithm Setup:}
    \begin{itemize}
      \item Use geometric epoch schedule with base $2.0$: epoch boundaries at $t = 2^k$ for $k=1,2,\ldots$ up to $T$.
      \item Two variants are compared:
        \begin{enumerate}
          \item \textbf{Theoretical excess risk:} Uses the bound $\varepsilon_m = C \cdot (dK + \ln(1/\delta)) / n$ where $d=10$ is the feature dimension, $K=5$ is the number of actions, $n$ is the total number of data points seen so far, $\delta=0.05$ is the confidence parameter, and $C=2$. The oracle fits $K$ separate $d$-dimensional ridge regressions (one per action), so the effective parameter count is $dK$; excess risk thus scales as O$(dK/n)$. 
          \item \textbf{Error-estimation excess risk:} Uses \texttt{ExcessRiskClassic} with:
            \begin{itemize}
              \item Linear regression model $f: (x, a) \mapsto \hat{r}(x, a)$ that maps encoded $(x, \text{one-hot}(a))$ vectors to predicted rewards.
              \item Definition set $S_{\text{def}}$: all cumulative data from previous epochs.
              \item Error set $S_{\text{err}}$: data from the current epoch only.
              \item The model $f$ is trained on $S_{\text{def}}$ using the linear regression oracle (ridge regression) to match oracle predictions for observed $(x, a, r)$ triples.
              \item Excess risk bound computed via variational optimization with Hoeffding correction, confidence parameter $\delta=0.05$, max iterations $200$, with timeout and retry mechanisms for robustness.
            \end{itemize}
        \end{enumerate}
      \item Both variants compute exploration parameter $\gamma_m = c \sqrt{K / \max(\varepsilon_m, \epsilon)}$ where $c=1.0$, $K=5$ (number of actions), and $\varepsilon_m$ is the excess risk term (theoretical or error-estimated).
      \item FALCON uses $\gamma_m$ to balance exploration vs.\ exploitation in its action selection policy.
    \end{itemize}

  \item \textbf{Per-Trial Procedure} (run in parallel via multiprocessing):
    \begin{enumerate}
      \item Generate random $\{\theta_a\}_{a=0}^4$ and fix contexts $\{x_t\}_{t=1}^T$ and reward noise $\{\varepsilon_t\}_{t=1}^T$ for this trial.
      \item Initialize both FALCON variants (theoretical and error-estimation) with the same configuration:
        \begin{itemize}
          \item Feature dimension $d=10$, number of actions $K=5$, confidence $\delta=0.05$, ridge regularization $\lambda=1.0$.
        \end{itemize}
      \item For each round $t=1,\ldots,T$:
        \begin{itemize}
          \item Both algorithms observe the same context $x_t$.
          \item Each selects an action $a_t$ according to its current policy (which depends on its $\gamma_m$).
          \item Both receive the same reward $r_t = x_t^\top \theta_{a_t} + \varepsilon_t$.
          \item Both update their models and compute new excess risk terms at epoch boundaries.
        \end{itemize}
      \item Record for each round:
        \begin{itemize}
          \item Excess risk term $\varepsilon_m$ (theoretical or error-estimated).
          \item Cumulative regret: $\sum_{s=1}^t \left[\max_a x_s^\top \theta_a - r_s\right]$.
        \end{itemize}
    \end{enumerate}

  \item \textbf{Aggregation:}
    \begin{itemize}
      \item Run $M=10$ independent trials with different random seeds (different $\{\theta_a\}$, contexts, and noise).
      \item For each metric (excess risk and cumulative regret), compute:
        \begin{itemize}
          \item Mean across trials: $\bar{y}_t = \frac{1}{M}\sum_{i=1}^M y_{i,t}$.
          \item Standard error: $\text{SE}_t = \frac{\text{std}(y_{1:M,t})}{\sqrt{n}}$.
        \end{itemize}
      \item Plot 95\% confidence bands: $\bar{y}_t \pm 1.96 \cdot \text{SE}_t$.
    \end{itemize}

  \item \textbf{Visualization:}
    \begin{itemize}
      \item \textbf{Top panel:} Excess risk term $\varepsilon_m$ over rounds (log scale), comparing theoretical bound vs.\ error-estimation bound. Error-estimation typically yields tighter (smaller) excess risk estimates, leading to larger $\gamma_m$ and better exploration-exploitation balance.
      \item \textbf{Bottom panel:} Cumulative regret over rounds. Lower regret indicates better performance. The error-estimation variant typically achieves lower cumulative regret due to more accurate excess risk estimates.
      \item Both panels show mean curves with shaded 95\% confidence bands across $M = 10$ trials.
    \end{itemize}

  \item \textbf{Implementation notes:}
    \begin{itemize}
      \item Trials are run in parallel using Python's \texttt{ProcessPoolExecutor} for efficiency.
      \item Error-estimation computations include per-epoch timeouts (60 seconds) and retry mechanisms (up to 2 retries with random reinitialization) to handle optimization failures gracefully, falling back to conservative theoretical bounds when needed.
      \item Results are saved to \texttt{falcon\_multi\_trial\_results\_multiprocess.pkl} and visualized via \texttt{analyze\_falcon\_comparison.py}.
    \end{itemize}
\end{enumerate}

\section{Details for proofs}
\label{sec:local_proof}
\subsection{Replacing $\max$ with $\sup$ for infinite $\calH$}
Formally, in the case of infinite classes, \Cref{thm:max-error-estimation-result} should be written with $\sup$ statements instead of $\max$. We restate this version here and prove it below.
\begin{theorem}[Data-Driven Upper Bound on Sup Error]
\label{thm:sup-error-estimation-result}
    Let $\calH$ be an infinite, continuous class (with continuous estimates and errors). Suppose \Cref{ass:base-error-estimate} holds. Let $\hatxi$ denote our estimated upper bound on the supremum of the errors, $\hatxi(S_{\err},\delta) := \sup_{h\in\calH}\hatu(S_{\err}, h,\delta)$. Then for any $\delta\in(0,1)$, we have that \eqref{eq:max-error-guarantee} and \eqref{eq:max-error-guarantee-wo-conditioning} hold.\
    \begin{equation}
    \label{eq:sup-error-guarantee}
        \Pr_{S_{\err}\sim\calD}\Big(\hatxi(S_{\err},\delta) \geq \sup_{h\in\calH}e_h\Big|\calE(s)\Big)\geq 1-\delta
    \end{equation}
    \begin{equation}
    \label{eq:sup-error-guarantee-wo-conditioning}
    \Pr_{(S_{\est},S_{\err})}\Big(\hatxi(S_{\err},\delta) \geq \sup_{h\in\calH}e_h\Big)\geq 1-\delta
    \end{equation}
\end{theorem}

\begin{proof}
    Define $E:=\sup_{h\in\calH}e_h$ and note that for every $\varepsilon > 0$, there exist $h_\varepsilon \in \calH$ such that $e_{h_\varepsilon} > E - \varepsilon$. Now, note
\begin{align*}
\Pr_{S{\err}\sim\calD}\Big(\hatxi(S_{\err},\delta) \geq E - \varepsilon \Big|\calE(s)\Big) & \ge\Pr_{S{\err}\sim\calD}\Big(\hatxi(S_{\err},\delta) \geq e_{h_\varepsilon} \Big|\calE(s)\Big)\\
& \geq \Pr_{S{\err}\sim\calD}\Big(\hatu(S_{\err},h_\varepsilon, \delta) \geq e_{h_\varepsilon} \Big|\calE(s)\Big) \geq 1-\delta
\end{align*}
Now, taking the sequence $\varepsilon_k = 1/k$, and the limit as $k \to \infty$, we have that \eqref{eq:sup-error-guarantee} holds. By the tower law, we also have that \eqref{eq:sup-error-guarantee-wo-conditioning} holds.
\end{proof}

\subsection{Proof of \Cref{cor:localize}}
Suppose \Cref{ass:specific-instance-error-estimate} holds, and consider the event $\event^*(S_{\err})=\{e_{h^*} \le \hatu(S_{\err}, h^*, \delta)\}$, and a class $\calG(S_\err) \subseteq \calH$ such that $\event^*(S_\err) \subseteq \{h^* \in \calG(S_\err)\}$. We then have
\begin{equation}
    \label{eq:localization-in-hp-event}
        \event^*(S_{\err}) \implies e_{h^*} \stackrel{(i)}{\leq} \hatu(S_{\err},h^*,\delta) \stackrel{(ii)}{\leq} \max_{h\in\calG(S_\err)}\hatu(S_{\err},h,\delta)
    \end{equation}
Here (i) follows from the definition of $\event^*(S_\err)$, and (ii) follows from the condition placed on $\calG(S_\err)$. We now use \eqref{eq:localization-in-hp-event} to show that
\begin{align*}
\label{eq:localized-probability-argument}
\Pr \left(e_{h^*} \le \max_{h \in \calG(S_\err)} \hatu(S_\err, h, \delta)|\calE(s)\right) &\stackrel{(i)}{\geq} \Pr_{S{\err}\sim\calD}(\event^*(S_\err)|\calE(s))\\
&\stackrel{(ii)}{=} \Pr_{S{\err}\sim\calD}\Big(\hatu(S_{\err},h^*,\delta) \geq e_{h^*}\Big|\calE(s)\Big) \stackrel{(iii)}{\geq} 1-\delta
\end{align*}
Here (i) holds by \eqref{eq:localization-in-hp-event}, (ii) holds by the definition of $\event^*(S_{\err})$, and (iii) holds by \Cref{ass:specific-instance-error-estimate}. Furthermore, if we can construct a sequence of classes $\{\calH_k(S_\err)\}_{k=0}^\infty \subseteq \calH$ where for each $k$, $\event^*(S_\err) \subseteq \{h^* \in \calH_k(S_\err)\}$, we can consider $\calH_\infty$, and by \Cref{ass:specific-instance-error-estimate}, $\Pr\left(e_{h^*} \le \max_{h \in \cap_k \calH_k(S_\err)} \hatu(S_\err, h, \delta)\right) \ge 1 - \delta$. \\

We construct $\calH_k$ as follows. Let $\calH_0(S_\err) = \calH$. Now define (as in \eqref{eq:localizers_def})
\begin{equation}
\label{eq:localizers}
    \hatxi_k(S_\err, \delta) :=  \max_{h \in \calH_k(S_\err)}\hatu(S_\err, h, \delta),\;\; \calH_{k+1}(S_\err) := \left\{h \in \calH| -\estimate_{h} \le \hatxi_{k}(S_\err, \delta) -c\right\}
    \end{equation}
We claim $\event^*(S_\err) \subseteq \{h^* \in \calH_k(S_\err)\}$ for all $k \in \mathbb{N}$. To show this we assume $\event^*(S_\err)$ holds and deduce $h^* \in \calH_k(S_\err) \forall k$ by induction. First note that this holds true for $k = 0$ since $h^* \in \calH_0(S_\err) = \calH$. Now, assume $h^* \in \calH_k(S_\err)$. Recall that $e_{h^*} = \theta_{h^*} - \estimate_{h^*} \ge c - \estimate_{h^*}$ by \Cref{ass:h*bound}. Thus, by definition of $\calH_{k+1}(S_\err)$, we have $h^* \in \calH_{k+1}(S_\err)$. 

\section{Additional details for \Cref{sec:ci-multiple-mean}}
\label{app:ee_example_details}
In this section, we show how to formally convert the example in \Cref{sec:intro} into the setting described in \Cref{sec:ci-multiple-mean}. Recall the goal of constructing simultaneously valid CIs on average test scores for subgroups (e.g.``female students", ``fourth-graders", etc.) indexed by $\calH$.\footnote{With mild abuse of notation, we let any $h\in\calH$ denote both the subgroup index and subgroup itself.} More formally, let $(X,T)\in\calX\times[-1,1]$ be a pair of random variables corresponding to a randomly sampled student's feature vector and their test score respectively. The average test score for any subgroup $h$ is then given by $\theta_h$ in \eqref{eq:avg-test-score-estimand}.
\begin{equation}
\label{eq:avg-test-score-estimand}
\begin{aligned}
    &\theta_h := \E[T|X\in h] = \E[Y(h)],\\ 
    & Y(h) := \frac{T \cdot \mathbf{1}(X\in h)}{p_h}, \; p_h := \Pr(X\in h). 
\end{aligned}
\end{equation}
For ease of exposition, we assume the probability ($p_h$) of a random student belonging to subgroup $h$ is known to us and is lower bounded by $1/M$ for some $M\in [1,\infty)$.\footnote{We can always estimate $p_h$ from $S_{\est}$ and reset $\calH \leftarrow \{h\in\calH| p_h \geq 1/M\}$ -- \Cref{cor:ci-multiple-mean} would still hold. The remarks would also hold if cross-fitting \cite{newey2018cross} were used to account for unknown $p_h$ in estimates from $S_{\est}$ and $S_{\err}$.}

Although the errors in this setting, unlike in \Cref{sec:ee_framework}, are normalized, we still observe that $\hatu$ inherits the distribution of $e$. We formalize this in the following remark.

\begin{remark}[Asymptotic Distribution of $\hatu$] \label{rmk:hatu-dist-multimean} As $|\bar{S}|\rightarrow \infty$, we have $\hatu(S_{\err},h,\delta)$ converges in distribution to $|\mathcal{N}(0,2)|+z_{\delta/2}$. Note that in this setting, $e_h$ converges in distribution to $|\mathcal{N}(0,1)|$.
\end{remark}
\begin{proof}
\begin{equation}
    \label{eq:proving-remark-on-distribution-of-hatu-in-multi-mean}
\begin{aligned}&\hatu(S_{\err},h,\delta)=\Bigg|\frac{\sqrt{n}((\estimate_{h}-\theta_h)+(\theta_h - \estimate_{\err,h})}{\hat{\sigma}_{h}}\Bigg| + \frac{\hat{\sigma}_{\err,h} z_{\delta/2}}{\hat{\sigma}_{h}}\\
    & \rightarrow |\mathcal{N}(0,2)| + z_{\delta/2}.
        \end{aligned}
    \end{equation}
From CLT and $S_{\est}\independent S_{\err}$, we get $\frac{\sqrt{n}(\estimate_{h}-\theta_h)}{\hat{\sigma}_{h}}, \frac{\sqrt{n}(\theta_h - \estimate_{\err,h})}{\hat{\sigma}_{\err,h}}$ (weakly) converge to i.i.d. $N(0,1)$ distributions. From law of large numbers, we have $\frac{\hat{\sigma}_{\err,h}}{\hat{\sigma}_{h}}\rightarrow 1$ (almost surely). Hence from Slutsky's lemma and \eqref{eq:asymptotic-hatu-for-multimean}, we get \eqref{eq:proving-remark-on-distribution-of-hatu-in-multi-mean}.
\end{proof}


\section{Multiple Hypothesis Testing}
\label{sec:multiple-hyps}
In hypothesis testing, we seek to compare a real-valued estimand of interest against a specified threshold. The null hypothesis claims the estimand is equal to the threshold, and the alternative hypothesis claims the estimand is greater than the threshold.\footnote{This choice of alternative hypothesis is without loss of generality. One may also consider an alternative hypothesis where the estimand is less than the threshold, or the estimand is not equal to the threshold.} Given a set $\calH$ of hypotheses, a standard goal is to reject as many null hypotheses as possible while controlling the family-wise error rate (FWER) to be at most some $\delta \in (0,1)$.\footnote{FWER is the probability of incorrectly rejecting at least one null hypothesis.} A popular approach to this problem is the Bonferroni correction, which can be seen as a generalization of the union bound in that it allows flexibility in assigning weights to different hypotheses. In this section, we use error-estimation to provide a data-driven alternative to the weighted Bonferroni correction. 

We convert the setup from \Cref{sec:ci-multiple-mean} into a multiple-hypothesis testing problem. Suppose, for every $h\in\calH$, the null hypothesis is $\theta_h= 0$\footnote{Without loss of generality, the threshold can be zero (by appropriately adjusting the estimand).} and the alternative hypothesis is $\theta_h> 0$. We reuse $\hatu$ constructed in \eqref{eq:asymptotic-hatu-for-multimean}, and introduce additional weight functions $\iota$ and $b$ over the class of hypotheses. Here $\iota$ is an indicator function allowing us to select which hypotheses to include in the testing procedure, and $b$ is a weighting function that allows us to adjust the relative importance for rejecting certain hypotheses in the error estimation process. These functions introduce flexibility in our approach, allowing us to use prior knowledge to focus on a specific subset or emphasize certain hypotheses.

\begin{corollary}
\label{cor:multiple-testing}
Consider the setting in \Cref{cor:ci-multiple-mean}. Let $w=(\iota,b)$ be any pair of weight functions such that  $w\independent S_{\err}$ -- where $\iota:\calH \rightarrow\{0,1\}$ and $b:\calH \rightarrow (0,\infty)$. We apply error estimation to bound the maximum weighted error, and the flexibility to select $w$ (using data in $S_{\est}$) should ideally allow us to maximize the number of null hypotheses rejected. Let $\hatxi_w(S_{\err},\delta):=\max_{\{h\in\calH|\iota(h)=1\}} b(h)\hatu(S_{\err},h,\delta)$. We now construct a set $\calH_r$ (see \eqref{eq:define-rejection-set}) -- such that for all $h\in\calH_r$, we reject the null hypothesis $\theta_h=0$ in favor of the alternative hypothesis $\theta_h>0$. We then have that 
\begin{align}
    &\calH_r:=\bigg\{h\in\calH\bigg| \iota(h)=1, \estimate_{h} > \frac{1}{b(h)}\frac{\hat{\sigma}_{\est,h}}{\sqrt{n}}\hatxi_w(S_{\err},\delta) \bigg\},\label{eq:define-rejection-set}\\
    &\Pr_{\bar{S}}\Big(\forall h\in\calH_r,\; \theta_h > 0\Big)\geq 1-\delta. \label{eq:multiple-testing-guarantee}
\end{align}
\end{corollary}
\begin{proof}
    Following the argument in \Cref{cor:ci-multiple-mean} (using \Cref{thm:max-error-estimation-result}), we have:
    \begin{equation}
        \begin{aligned}
            &\Pr_{\bar{S}}\Big(\max_{\{h\in\calH|\iota(h)=1\}} b(h)e_{h}\leq  \hatxi_w(S_{\err},\delta)\Big)\geq 1-\delta\\
            \equiv&\Pr_{\bar{S}}\Big(\forall h\in\calH \text{ s.t. } \iota(h)=1, |\estimate_{h} - \theta_h|\leq  \frac{1}{b(h)}\frac{\hat{\sigma}_{\est,h}}{\sqrt{n}}\hatxi_w(S_{\err},\delta)\Big)\geq 1-\delta\\
            \implies&\Pr_{\bar{S}}\Big(\forall h\in\calH \text{ s.t. } \iota(h)=1, \theta_h\geq \estimate_{h} - \frac{1}{b(h)}\frac{\hat{\sigma}_{\est,h}}{\sqrt{n}}\hatxi_w(S_{\err},\delta)\Big)\geq 1-\delta\\
            \implies&\Pr_{\bar{S}}\Big(\forall h\in\calH_r,\; \theta_h > 0\Big)\geq 1-\delta.
        \end{aligned}
    \end{equation}
\end{proof}
Note that FWER for \Cref{cor:multiple-testing} is at most $\delta$ for any choice of $w$. Now let us briefly discuss the choices for $w=(\iota,b)$. A simple choice is to let $b\equiv\iota\equiv 1$, this choice makes \Cref{cor:multiple-testing} a direct application of \Cref{cor:ci-multiple-mean}. However, while keeping FWER less than $\delta$, a more intelligent choice for $w$ will allow us to discover more $h$ for which the alternative hypothesis is true (larger $\calH_r$). 
Note that even for rejecting the single null hypothesis $\theta_h=0$ with dataset $S_{\est}$, we would require $\estimate_{h} > \hat{\sigma}_{\est,h}z_{\delta}/\sqrt{n}$. Hence for multiple testing, we only need to focus our attention on $h$ satisfying this condition. It would thus be reasonable to choose $\iota(h)=\mathbf{1}( \estimate_{h} > \hat{\sigma}_{\est,h}z_{\delta}/\sqrt{n})$, ensuring $\hatxi_w(S_{\err},\delta)$ is not larger than necessary. Using $b(h)=\hat{\sigma}_{\est,h}\min(1,1/|\estimate_{h}|)$ is also a reasonable choice -- this makes the tests in \eqref{eq:define-rejection-set} more comparable -- allowing for $\calH_r$ to be much larger when possible.
If we are willing to incur a larger FWER in exchange for more nulls rejected, we can set $\iota(h)$ as zero more often and extrapolate the test in \eqref{eq:define-rejection-set} to such $h$. Such approaches could allow us to tackle objectives like controlling the false discovery rate (FDR). Analyzing the choices of $w$ and its various implications is beyond the scope of this paper.

\section{Improved performance via cross-fitting}
\label{app:crossfit}
Our basic error-estimation construction uses a single split: a defining dataset $S_{\est}$ to form the estimates $\{\hat\theta_h\}_{h\in\calH}$ (and the induced class $\calE(s)$), and an independent error-estimation dataset $S_{\err}$ to compute the surrogate bounds $\hatu(S_{\err},h,\delta)$. This split can inflate the resulting uniform bound, since each direction uses only half the data for estimation.

A simple mitigation is \emph{two-way dataset switching} (a $K{=}2$ cross-fitting variant). Let the full sample be split into two folds $S^{(1)}$ and $S^{(2)}$ (e.g., equal size). We run the method in both directions:
(i) define $\{\hat\theta_h\}$ using $S^{(1)}$ and compute the uniform bound using $S^{(2)}$, yielding
\[
\hat\xi^{1\to 2}(\lambda)\;:=\;\sup_{h\in\calH}\hatu\!\left(S^{(2)},h,\lambda\right);
\]
(ii) swap the roles of the folds, yielding
\[
\hat\xi^{2\to 1}(\lambda)\;:=\;\sup_{h\in\calH}\hatu\!\left(S^{(1)},h,\lambda\right).
\]
We then report the \emph{switched} bound
\[
\hat\xi^{\min}(\delta)\;:=\;\min\Big\{\hat\xi^{1\to 2}(\delta/2),\;\hat\xi^{2\to 1}(\delta/2)\Big\}.
\]

\begin{figure}[t]
\centering
\includegraphics[width=0.5\textwidth]{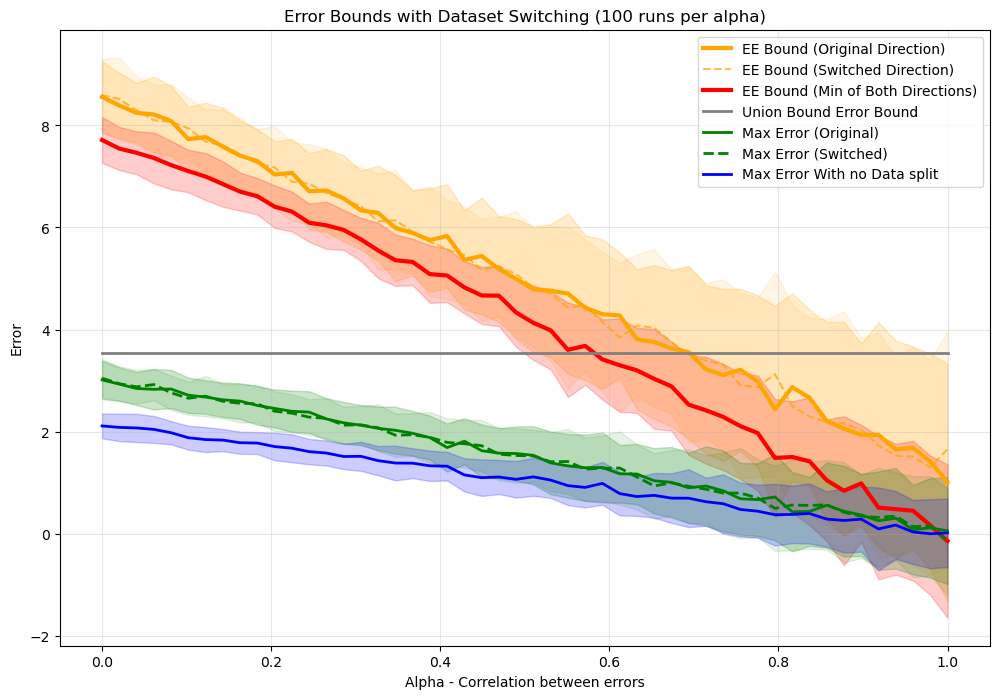}
\caption{Effect of two-way dataset switching in the correlated finite-class simulation (100 runs per $\alpha$).
Orange: error-estimation bound using the original split direction; dashed orange: the same procedure after swapping the roles of $S_{\est}$ and $S_{\err}$; red: the switched bound $\hat\xi^{\min}(\delta)$.
Green curves show the corresponding realized max errors under each direction, and the blue curve shows the max error computed without a split (oracle baseline). Switching and taking the minimum substantially tightens the bound, often mitigating a large fraction of the split-induced looseness.}
\label{fig:dataset_switching}
\end{figure}

Empirically, Figure~\ref{fig:dataset_switching} shows three regimes: (i) the original direction, (ii) the switched direction, and (iii) taking the minimum of the two valid bounds (with $\delta/2$ in each direction).
In this simulation, the min-of-two procedure often reduces conservatism by a sizable constant factor, bringing the bound closer to the no-split oracle baseline.

\paragraph{Beyond two folds.}
More generally, one can repeat this idea over multiple random resplits (or $K$ folds), compute $m$ valid bounds at confidence levels $\delta/m$, and take their intersection (equivalently, the minimum of the $m$ upper bounds). At some point the additional Bonferroni cost from using $\delta/m$ will outweigh the benefit of intersecting, so selecting the number of resplits is a practical tradeoff; understanding this tradeoff more sharply is an interesting direction for future work.

%% file: icml2026/appendix_bandits_pipeline.tex
\section{Contextual Bandit Pipeline with Error Estimation}
\label{sec:cbs}

In this section, we describe a modular contextual bandit pipeline that leverages error estimation. We begin by introducing the standard stochastic contextual bandit setting, defined by a context space $\calX$, a finite set of arms $\A = [K]$, and a fixed but unknown distribution $\calD$ over contexts and arm rewards.\footnote{Here $[n]$ denotes the set $\{1,2,\dots,n\}$.} A contextual bandit algorithm (learner) interacts with samples from this unknown distribution $\calD$ over a sequence of 
rounds. At each round $t$, a context $x_t$ and a reward vector $r_t \in [0,1]^{\calA}$ are sampled from the joint distribution $\calD$. Initially, only the context $x_t$ is revealed to the learner. To make a decision, the learner utilizes data from prior rounds to construct an ``exploration policy" $p_t$ -- which is a probability kernel mapping $\calX\times\calA$ to $[0,1]$. The learner then chooses an arm $a_t$ sampled from the distribution $p_t(\cdot|x_t)$ and observes the corresponding reward $r_t(a_t)$, this data is later used for constructing exploration policies in subsequent rounds.

\textbf{Key concepts.} Before defining our objectives, we introduce several key concepts. A reward model $f$ maps $\calX\times\A$ to $[0,1]$, with $f^*(x,a):=\E_D[r_t(a)|x_t=x]$ denoting the true conditional expected reward model. A (deterministic) policy $\pi$ is a map from $\calX$ to $\A$, and a randomized policy $p$ is a probability kernel mapping $\calX\times\A$ to $[0,1]$. For a given model $f$ and a randomized policy $p$, we denote the expected instantaneous reward of $p$ with respect to model $f$ as $R_f(p)$ -- formally defined in \eqref{eq:def-reward-and-cover}. We denote the true expected reward $R_{f^*}(p)$ as $R(p)$ to simplify notation when there is no risk of confusion. We also let $V(p,\pi)$ denote the ``cover" of policy $\pi$ under data collected by a (randomized) policy $p$, which is formally defined in \eqref{eq:def-reward-and-cover}  -- it is proportional to the variance of evaluating the policy $\pi$ using data collected by the policy $p$. Lower values of the cover indicate that data collected under $p$ provides more precise and reliable estimates for the performance of $\pi$.
\begin{equation}
\label{eq:def-reward-and-cover}
    \begin{aligned}
        R_f(p) := \E_{(x,r) \sim D}\E_{a\sim p(\cdot|x)}[f(x, a)],\quad \text{and } V(p,\pi):=\E_{(x,r)\sim D, a\sim \pi(\cdot|x)}\bigg[\frac{\pi(a|x)}{p(a|x)}\bigg]. 
    \end{aligned}
\end{equation}
Our learner (bandit algorithm) is given a policy class $\Pi$. Let $\pi^*$ denote the optimal policy in the class $\Pi$, that is $\pi^*\in\arg\max_{\pi\in\Pi} R(\pi)$. 



\textbf{Objective.} Contextual bandit algorithms run in epochs (batches). At the start of each epoch, a new exploration (data-collection) policy is computed and is used to select arms for each round in that epoch. Consider an epoch that starts at round $\tau+1$. To construct an exploration policy $p$ for this epoch, we leverage dataset $S$ corresponding to the first $\tau$ rounds. Arguably, designing a contextual bandit algorithm boils down to designing an algorithm that uses data from prior epochs to construct the new exploration policy $p$. Hence, to simplify our analysis and focus on algorithmic insights, we study an objective that is solely a function of $p$. In particular, our objective is to construct $p$ in order to minimize the ``optimal cover", defined as $V(p,\pi^*)$. The optimal cover is proportional to the variance of evaluating the optimal policy and has been shown to be an effective surrogate objective for pure exploration \citep[see discussions in][]{krishnamurthy2023proportional, maurer2009empirical}.\footnote{At a high-level, minimizing the variance of evaluating the optimal policy allows variance-penalized policy learning algorithms to ensure better guarantees \citep{xu2020towards}. Hence minimizing the optimal cover is seen as a surrogate objective for pure exploration.}  Furthermore, although there can be some trade-off between minimizing $V(p,\pi^*)$ and maximizing $R(p)$, minimizing the optimal cover encourages behavior similar to the optimal policy and thus naturally results in exploration policies with high expected reward.

\textbf{Input.} The input dataset $S$ corresponds to data from the first $\tau$ rounds, collected under a doubling epoch schedule.\footnote{That is, last $\tau/2$ rounds are collected under a single exploration policy. For any $t,t'\in (\tau/2,\tau]$, we have $p_t=p_{t'}$.} For each round $t\in [\tau]$, input $S$ contains: (i) the observed context $x_t$, (ii) the selected arm $a_t$, (iii) the observed reward $r_t(a_t)$, (iv) the exploration policy $p_t$ used at round $t$, (v) a high-probability upper bound $\alpha_t$ on the optimal cover ($\alpha_t\geq V(p_t,\pi^*)$), and (vi) a high-probability upper bound $M_t$ on the inverse probability of sampling the optimal policy under the exploration policy $p_t$, i.e. such that $1/p_t(\pi^*(x)|x)\leq M_t$. All bounds provided in dataset $S$ must simultaneously hold with probability at least $1-\deltaprior$, for some $\deltaprior\in [0,1)$. To simplify notation, we allow $t\in[\tau]$ to denote both the round index and all the corresponding data in dataset $S$. In \eqref{eq:def-tilde-Pi}, we define $\Tilde{\Pi}$, a subclass of possibly optimal policies. 
\begin{equation}
\label{eq:def-tilde-Pi}
    \Tilde{\Pi}:=\{\pi\in\Pi|p_t(\pi(x)|x)\geq 1/M_t, \forall x\in\calX,t\in[\tau] \}.
\end{equation}

\subsection{Contextual Bandit Pipeline}

\begin{algorithm}[h!]
  \caption{Contextual Bandit Pipeline: updating exploration policy at the end of an epoch}
  \label{alg:GenRAPR}
  \textbf{Inputs:} Dataset $S$ from first $\tau$ rounds. Optimal cover guarantee for prior rounds $\{\alpha_t| V(p_t,\pi^*)\leq \alpha_t, \forall t\in[\tau] \}$. Guarantee on sampling probability of $\pi^*$ in prior rounds, $\{M_t|p_t(\pi^*(x)|x)\geq 1/M_t, \forall x\in\calX,t\in[\tau] \}$. Let $\Tilde{\Pi}:=\{\pi\in\Pi|p_t(\pi(x)|x)\geq 1/M_t, \forall x\in\calX,t\in[\tau] \}$.\\ 
  \textbf{Parameters:} Confidence parameter $\delta\in(0,1)$. Error estimation fraction $\lambda\in(0,1/2)$. Proportional response threshold $\beta_{\max}\in(0,1)$. 
  \begin{algorithmic}[1] 
  \STATE Split $S$: Let $S_{\est}$ be data from first $\lfloor(1-\lambda)\tau\rfloor$ rounds, and $S_{\err}$ be remaining data, with exploration policy $p_\err$ and optimal cover bound $\alpha_\err \ge V(p_\err, \pi^*)$. We further split $S_{\err}$ into $(S_{\err, \con}, S_{\err, \elim}, S_{\err, B})$.
  \STATE Estimate $(\hatR_{\elim},\hat{\pi}_{\elim}) \leftarrow \BlackBoxPolicyEvaluator(S_{\est})$. \COMMENT{For any policy $\pi$, $\hatR_{\elim}(\pi)$ is an estimate of $R(\pi)$ and can be estimated using any procedure. $\hatpi_{\elim}\in\Tilde{\Pi}$ is ideally chosen to maximize $\hatR_{\elim}(\cdot)$. These outputs are used for contextual arm elimination.}
  \STATE Estimate $\hatU_\elim \leftarrow\CBElimErrorEstimator (S_{\est},S_{\err, \elim},\hatR_{\elim},\hat{\pi}_{\elim},\delta, \alpha_\err)$ such that \eqref{eq:policy-based-error-estimation} holds.
  \begin{equation}
  \label{eq:policy-based-error-estimation}
      \begin{aligned}
          &\Pr_{S_{\err, \elim} \sim \calD}\Big( \hatR_{\elim}(\hat{\pi}_{\elim}) - \hatR_{\elim}(\pi^*)\leq \hatU_\elim \Big)\geq 1-\delta/3.
      \end{aligned}
  \end{equation}
  \STATE Compute $\hatg \leftarrow \ArmEliminator(\hatR_{\elim},\hat{\pi}_{\elim},\hatU_\elim,S)$ such that \eqref{eq:ideal-arm-eliminator-guarantee} holds.
  \begin{equation}
  \label{eq:ideal-arm-eliminator-guarantee}
      \hatg (x) \supseteq \bigg\{ \pi(x)|\pi\in\Pi, \hatR_{\elim}(\hat{\pi}_{\elim}) - \hatR_{\elim}(\pi)\leq \hatU_\elim \bigg\} \forall x.
  \end{equation}
  \STATE Estimate $(\hatf,\hatpi_{\con}) \leftarrow \BlackBoxRegressor(S_{\est}, S_{\err, \elim})$.
  \COMMENT{Here $\hatf$ is an estimate of $f^*$ and can be estimated using any class/procedure. $\hatpi_{\con}\in\Tilde{\Pi}$ is ideally chosen to maximize reward, $R_{\hatf}(\hatpi_{\con})$. These outputs are used to construct conformal arm sets.}
  \STATE Estimate $\hatU_\con,\leftarrow\CBConErrorEstimator((S_{\est},S_{\err, \elim}),S_{\err, \con},\hatf,\hat{\pi}_{\con},\hatR_{\est},\hat{\pi}_{\elim},\delta, \alpha_\err)$ such that \eqref{eq:reward-based-error-estimation} holds.
  \begin{equation}
  \label{eq:reward-based-error-estimation}
      \begin{aligned}
      &\Pr_{S_{\err, \con} \sim \calD}\Big(R_{\hatf}(\hatpi_{\con}) - R_{\hatf}(\pi^*) \leq \hatU_\con \Big)\geq 1-\delta/3,
      \end{aligned}
  \end{equation}
    \STATE Define Conformal Arm Sets \eqref{eq:CAS-based-on-error-estimation}. Let $\Unif_{\zeta}(\cdot|x)$ denote the uniform distribution over arms in $C(x,\zeta)$.
  \begin{equation}
  \label{eq:CAS-based-on-error-estimation}
      C(x, \zeta) := \Bigg\{a \in \hatg(x): 
    \hatf (x, \hatpi_{\con}(x)) - \hatf (x,a)\leq \frac{\hatU_{\con}}{\zeta}
    \Bigg\}.
  \end{equation}
  \STATE Select risk adjustment parameter $\eta\in[1,K]$. \COMMENT{Attempting to minimizes the optimal cover upper bound in \Cref{thm:cbpipelinemain}.}
  \STATE Return the next exploration policy $p_{\tau +1}$,  a guarantee on its sampling probability of $\pi^*$, $M_{\tau + 1}$, and a bound on its optimal cover $\alpha_{\tau + 1}$ as defined in \eqref{eq:define_pma_genRAPR} . 
    \begin{equation}
    \label{eq:define_pma_genRAPR}
    \begin{aligned}
        &p_{\tau + 1} (a|x) = (1-\beta_{\max})\Unif_{\beta_{\max}/\eta}(a|x)+ \int_{0}^{\beta_{\max}}\Unif_{\beta/\eta}(a|x)\mbox{d}\beta.\\
        &M_{\tau + 1} =  \frac{\max_x |\hatg(x)|}{\eta\hatU}\\
        &\alpha_{\tau +1} = \E_{x\sim S_{\err, B}}\Bigg[\frac{|C(x, \propThreshold/\eta)|}{1-\propThreshold + \propThreshold \frac{|C(x, \propThreshold/\eta)|}{|\hatg(x)|} }\Bigg] + \frac{\max_x |\hatg(x)|}{\eta} + \sqrt{\frac{\max_x | \hatg(x)|\log(3/\delta)}{\beta_{\max}|S_{\err, B}|}}\\
    \end{aligned}
    \end{equation}
  \end{algorithmic}
\end{algorithm}

\begin{figure}[h]
    \centering
    \begin{tikzpicture}
        \definecolor{collection1}{RGB}{173, 216, 230} 

        \fill[collection1] (0,0) rectangle (15,0.5); 
        
        \draw[very thick, white] (0.4845,0) -- (0.4845,0.5); 
        \draw[very thick, white] (1.452,0) -- (1.452,0.5); 
        \draw[very thick, white] (3.387,0) -- (3.387,0.5); 
        \draw[very thick, white, dotted] (4.677,0) -- (4.677,0.5); 
        \draw[very thick, white, dotted] (5.9685,0) -- (5.9685,0.5); 
        \draw[very thick, white] (7.2585,0) -- (7.2585,0.5); 
        
        \draw [decorate,decoration={brace,amplitude=10pt,mirror}] (0,-0.2) -- (3.387,-0.2) node[midway,yshift=-15pt] {\small \textbf{$S_\est$}};
        \draw [decorate,decoration={brace,amplitude=10pt,mirror}] (3.387,-0.2) -- (7.2585,-0.2) node[midway,yshift=-15pt] {\small \textbf{$S_\err$}};
        \draw [decorate,decoration={brace,amplitude=10pt,mirror}] (7.2585,-0.2) -- (15,-0.2) node[midway,yshift=-15pt] {\small Next round explored with $p_\err$};

        \draw [decorate,decoration={brace,amplitude=5pt}] (3.387,0.5) -- (4.677,0.5) node[midway,yshift=10pt] {\small \textbf{$S_{\err, \elim}$}};
        \draw [decorate,decoration={brace,amplitude=5pt}] (4.677,0.5) -- (5.9685,0.5) node[midway,yshift=10pt] {\small \textbf{$S_{\err, \con}$}};
        \draw [decorate,decoration={brace,amplitude=5pt}] (5.9685,0.5) -- (7.2585,0.5) node[midway,yshift=10pt] {\small \textbf{$S_{\err, B}$}};

    \end{tikzpicture}
    \caption{Visualization of data allocation in epoch $\tau$, with $\lambda = 1/2$}
    \label{fig:data-allocation}
\end{figure}

This section describes our contextual bandit pipeline as detailed in \Cref{alg:GenRAPR}. To facilitate a truly modular contextual bandit pipeline, we employ ``black box" subroutines without specific requirements on output quality whenever possible. This allows for the use of a wide range of heuristic subroutines that perform well in practice but may lack strong theoretical guarantees (e.g., methods based on deep learning or fine-tuning). To create bounds on the suboptimality of the performance of these black box subroutines, we use the error estimation theory derived above. Therefore, our algorithm begins by splitting the input dataset $S$ into $S_{\est}$ and $S_{\err}$,\footnote{ensuring that the rounds in $S_{\err}$ follow those in $S_{\est}$.} and then processes this data using several subroutines. As a result of the doubling epoch schedule, the $S_{\err}$ dataset is collected under a single exploration policy $p_{\err}$. We take as input a bound on the optimal cover of this exploration policy, $\alpha_\err \ge V(p_{\err}, \pi^*)$. The algorithm will be using error estimation theory in two separate instances. Additionally, we set off $S_{\err, B}$ to bound the optimal cover of the next exploration policy. Accordingly we split $S_\err$ into $S_{\err, \elim}$, $S_{\err, con}$, and $S_{\err, B}$.  (see \Cref{fig:data-allocation} for outline of data allocation in round $\tau$).\\

We begin by applying a black box subroutine ($\BlackBoxPolicyEvaluator$) to dataset $S_{\est}$ to estimate a policy evaluator $\hatR_{\elim}:\Pi \rightarrow [0,1]$, and construct policy $\hatpi_{\elim} = \argmax_{\pi\in\Tilde{\Pi}}\hatR_{\elim}(\pi)$. Ideally, $\hatR_{\elim}$ is equal to $R$. The closer these two policy evaluators are, the better the performance of downstream tasks.

To estimate a high-probability upper bound on the $\hatR_\elim(\hatpi_\elim) -\hatR_\elim(\pi^*)$ from the output of $\BlackBoxPolicyEvaluator$, we use a subroutine ($\CBElimErrorEstimator$) on the dataset $S_{\err, \elim}$. We require the output from this subroutine ($\hatU_\elim$) to satisfy \eqref{eq:policy-based-error-estimation} and discuss one approach (based on error estimation) to designing such a subroutine in \Cref{sec:cberrorestimation} using $S_{\err, \elim}$. With this error bound, we use subroutine ($\ArmEliminator$) to construct $\hatg:\calX\rightarrow 2^{\calA}$ satisfying \eqref{eq:ideal-arm-eliminator-guarantee}. Note that \eqref{eq:ideal-arm-eliminator-guarantee} is not a statistical requirement, and the challenge in designing such a subroutine is purely computational. A trivial way to satisfy \eqref{eq:ideal-arm-eliminator-guarantee} is to choose $\hatg(x)=\calA$ for all $x$. We describe a more non-trivial and implementable approach to designing such a subroutine in \Cref{sec:ArmEliminator}.

After having established a set of potentially optimal arms for every context, we apply another black box subroutine ($\BlackBoxRegressor$) to estimate a reward model $\hat{f}:\calX \times \calA \rightarrow [0,1]$ and construct a policy $\hatpi_{\con} = \argmax_{\pi\in\Tilde{\Pi}}R_{\hatf}(\pi)$. Ideally, $\hatf$ is equal to $f^*$.

 Once again, we describe an approach using error estimation to estimate a high probability upper bound on $R_{\hatf}(\hat{\pi}_{\con}) - R_{\hatf}(\pi^*)$, a measure of the quality of $\hatf$ as the output of $\BlackBoxRegressor$. We require the output ($\hatU_\con$) to satisfy \eqref{eq:reward-based-error-estimation} and design a subroutine ($\CBConErrorEstimator$) in \Cref{sec:cberrorestimation} using data $S_{\err, \con}$. 

Finally we can define the sets of arms we will explore in the next round. In  \eqref{eq:CAS-based-on-error-estimation} we construct conformal arm sets (CASs) for every context $x \in \calX$. These are arms covered by potentially optimal policies according to $\hatR_{\elim}$ (up to some error $\hatU_\elim$), which are estimated to have high rewards according to $\hatf$ (up to some error $\hatU_\con$). We then select a risk-adjustment parameter $\eta$, and construct our new exploration (data-collection) policy $p_{\tau + 1}$ \citep[see][for a detailed discussion on CASs and using them for designing exploration policies with low optimal cover]{krishnamurthy2023proportional}.\\

Finally, we can upper bound both the optimal cover and the maximum inverse probability of sampling the arm recommended by the optimal policy (see \eqref{eq:define_pma_genRAPR}). Here we use $S_{\err, B}$. See \Cref{thm:cbpipelinemain} for formal theoretical guarantees.

$\CBElimErrorEstimator$ \eqref{eq:policy-based-error-estimation}, $\CBConErrorEstimator$ \eqref{eq:reward-based-error-estimation} and the bound for $\alpha_{\tau + 1}$ \eqref{eq:define_pma_genRAPR} are the only subroutines in \Cref{alg:GenRAPR} with a statistical guarantee.

\begin{restatable}[]{theorem}{cbpipelinemain}
\label{thm:cbpipelinemain}
    Consider an input dataset $S$ corresponding to data from the first $\tau$ rounds, where all bounds provided in dataset $S$ hold simultaneously with probability at least $1-\deltaprior$. Suppose we run \Cref{alg:GenRAPR} on input $S$ with the following algorithmic parameters: (i) confidence parameter $\delta\in(0,1)$, (ii) error estimation fraction $\lambda\in(0,1)$, and (iii) proportional response threshold $\beta_{\max}\in(0,1)$. Let $p_{\tau + 1}$, $M_{\tau + 1}$, and $\alpha_{\tau + 1}$ be the exploration policy returned by \Cref{alg:GenRAPR}, and bounds on its minimum sampling probability of $\pi^*$ and on its optimal cover respectively, as constructed in \eqref{eq:define_pma_genRAPR}. Further suppose the outputs of $\CBElimErrorEstimator$, $\CBConErrorEstimator$ and $\ArmEliminator$ satisfy the expected guarantees. That is, $\hatU_{\elim},\hatU_{\con}$ satisfy the high-probability guarantees in \eqref{eq:policy-based-error-estimation} and \eqref{eq:reward-based-error-estimation} respectively. Moreover, $\hatg$ is constructed to satisfy \eqref{eq:ideal-arm-eliminator-guarantee}. With probability $1-\deltaprior-\delta$, in addition to the bounds in $S$ holding, we also have \eqref{eq:pipeline-guarantees} holds.
    \begin{equation}
    \label{eq:pipeline-guarantees}
        \begin{aligned}
            &\pi^*(x)\in \hatg(x),\forall x;\;\;\;  \max_{x}\frac{1}{p(\pi^*(x)|x)} \leq M_{\tau + 1} = \frac{\max_x |\hatg(x)|}{\eta\hatU_{\con}};\\ 
            &V(p_{\tau + 1},\pi^*)\leq \alpha_{\tau + 1} = \E_{x\sim S_{\err, B}}\Bigg[\frac{|C(x, \propThreshold/\eta)|}{1-\propThreshold + \propThreshold \frac{|C(x, \propThreshold/\eta)|}{|\hatg(x)|} }\Bigg] + \frac{\max_x |\hatg(x)|}{\eta} + \sqrt{\frac{\max_x | \hatg(x)|\log(3/\delta)}{\beta_{\max}|S_{\err, B}|}}.
        \end{aligned}
    \end{equation}  
\end{restatable}
Note that with an appropriate choice of $\eta$, the optimal cover bound from \Cref{thm:cbpipelinemain} primarily depends on the average size of the set $C(x, \propThreshold/\eta)$ -- which would be small when the $\CBErrorEstimator$ outputs ($\hatU_{\elim}$ and $\hatU_{\con}$) are small -- which in turn can only happen when outputs of the black box subroutines have fundamentally small errors.

\subsection{Error Estimation for Contextual Bandits}
\label{sec:cberrorestimation}
In this section, we apply error estimation theory to develop the two $\CBErrorEstimator$ subroutines, ensuring they meet the criteria specified in equations \eqref{eq:policy-based-error-estimation} and \eqref{eq:reward-based-error-estimation}. We present an overview of the terms used in \Cref{tab:ee_cb}. While $\CBElimErrorEstimator$ implements error estimation in a manner very similar to the application in \Cref{sec:excess-risk-error-estimation}, $\CBConErrorEstimator$ involves an extra step, as the "estimate" quantity is not directly accessible. We outline the methods and algorithms in this section, and provide detailed proofs in \Cref{app:ee}.

\begin{table}[ht]
\centering
\caption{Key Error Estimation Quantities for Arm Elimination and Conformal Arm sets}
\label{tab:ee_cb}
\resizebox{\textwidth}{!}{%
\begin{tabular}{>{\raggedright\arraybackslash}p{3cm} p{6.5cm} p{6.5cm}}
\toprule
 & \multicolumn{1}{c}{\textbf{Arm Elimination}} & \multicolumn{1}{c}{\textbf{Contextual Arm Sets}} \\
\midrule
\textbf{Estimand} & \(\theta^{\elim}(\pi) = R(\pi) - R(\hat{\pi}_{\elim})\) & \(\theta^{\con}(\pi) = R(\pi) - R(\hat{\pi}_{\con})\)  \\
\midrule
\textbf{\(S_{\est}\) Estimate} & \(\hat{\theta}^{\elim}_{\hatR}(\pi) = \hat{R}_{\elim} (\pi) - \hat{R}_{\elim}(\hat{\pi}_{\elim})\) & \(\hat{\theta}^{\con}_{\hatf, \err_X}(\pi) = \mathbb{E}_{x \sim S_{\err, \con}}[\hat{R}_{\hatf} (\pi) - \hat{R}_{\hatf}(\hat{\pi}_{\con})]\) \\
\midrule
\textbf{Error} & \(e_\pi' = \theta^{\elim}(\pi) - \hat{\theta}^{\elim}_{\hatR}(\pi)\) & \(e_\pi = \theta^{\con}(\pi) - \hat{\theta}^{\con}_{\hatf, \err_X}(\pi)\) \\
\midrule
\textbf{\(S_{\err}\) Estimate} & \(\hat{\theta}^{\elim}_{\err}(\pi) = \mathbb{E}_{(x,a,r) \sim S_{\err, \elim}}[\theta^{\elim}(\pi)]\) & \(\hat{\theta}^{\con}_{\err}(\pi) = \mathbb{E}_{(x,a,r) \sim S_{\err, \con}}[\theta^{\con}(\pi)]\)  \\
\bottomrule
\end{tabular}
}
\end{table}

\paragraph{Bounding error used in Arm Elimination \eqref{eq:policy-based-error-estimation}}

We begin with constructing a bound $\hatU_\elim$ to satisfy \eqref{eq:policy-based-error-estimation}. We define our estimand analogously to the supervised learning setting described in \Cref{sec:excess-risk-error-estimation},  $\theta^{\elim}(\pi) =  
R (\pi) - R(\hatpi_{\elim})$.\footnote{Note that while in the setting of Section \cref{sec:excess-risk-error-estimation}, $\calR$ was a measure of risk, in this setting, $R$ represents rewards. Thus, although in \Cref{sec:excess-risk-error-estimation}, we had $\theta_g = \calR(g_{\est}) - \calR(g)$, we now have $\theta^{\elim}(\pi) =  
R (\pi) - R(\hatpi_{\elim})$, i.e. the signs have switched.} This is a measure of sub-optimality of policy $\hatpi_{\elim}$ with respect to a particular candidate policy $\pi$. However, note that while the setting in \Cref{sec:excess-risk-error-estimation} was motivated by providing a bound on the estimand $\theta_{g^*}$, in this setting we want to bound our estimate, $-\hat{\theta}^{\elim}_{\hatR}(\pi^*)$, based on our policy evaluator $\hatR_{\elim}(\pi)$. We define our estimate and error as
\begin{equation}
\label{eq:cb_elim_ee-estimate_error}
    \hat{\theta}^{\elim}_{\hatR}(\pi) := \hat{R}_{\elim} (\pi) - \hat{R}_{\elim}(\hatpi_{\elim}), \quad e^{\elim}_{\pi} := \theta^{\elim}(\pi)  - \hat{\theta}^{\elim}_{\hatR}(\pi)
\end{equation}

To bound $-\hat{\theta}^{\elim}_{\hatR}(\pi^*)$ as needed in \eqref{eq:policy-based-error-estimation}, we use the \textbf{key insight} that $\theta^{\elim}(\pi^*) = R(\pi^*) - R(\hatpi_{\elim}) \ge 0$ and therefore $e^{\elim}_{\pi^*} \ge - \hat{\theta}^{\elim}_{\hatR}(\pi^*)$. This insight is useful for two reasons. One, it allows us to use $e^{\elim}_{\pi^*}$ as an upper bound on our quantity of interest (and hence leverage our error estimation method). Two, this allows us to satisfy \Cref{ass:h*bound}, and therefore use \Cref{cor:localize} to improve our bounds. Now, we use $S_{\err, \elim}$ to estimate $\theta^{\elim}(\pi) $, so we define \[\hat{\theta}^{\elim}_{\err}(\pi) := \mathbb{E}_{(x,a,r) \sim S_{\err, \elim}}[R (\pi) - R(\hatpi_{\elim})]\] to construct a quantity $\hatu_{\elim}$ to satisfy \Cref{ass:specific-instance-error-estimate}.
\begin{equation}
    \hatu_{\elim}(\pi, S_{\err, \elim}, \delta) := \sqrt{\frac{\log(1/\delta)}{|S_{\err, \elim}|}}\left(\sqrt{V(p_\err, \hatpi_{\elim}) } +  \sqrt{\alpha_\err}\right) + \hat{\theta}^{\elim}_{\err}(\pi) - \hat{\theta}^{\elim}_{\hatR}(\pi), \label{eq:hatu_elim_cb}
\end{equation}
This gives us the first bound $- \hat{\theta}^{\elim}_{\hatR}(\pi^*) \le e^{\elim}_{\pi^*} \le \hatxi_0^{\elim}(\delta) := \max_{\pi \in \Tilde{\Pi}} \hatu_{\elim}(\pi, S_{\err, \elim}, \delta)$. Now we apply localization as in \Cref{cor:localize} with $c = 0$, set $\Pi_0^{\elim} = \tilde{\Pi}$, and for $k \ge 1$,
\[\Pi^{\elim}_k := \left\{\pi \in \Pi^{\elim}_{k-1}|- \hat{\theta}^{\elim}_{\hatR}(\pi) \le \hatxi_{k}^{\elim}( \delta) \right\}, \text{ where } \hatxi_k^{\elim}(\delta) := \max_{\pi \in \Pi^{\elim}_{k-1}} \hatu_{\elim}(\pi, S_{\err, \elim}, \delta) \]
By \Cref{cor:localize}, we have \eqref{eq:elim_error} holds.
\begin{equation}
\label{eq:elim_error}
    \Pr_{S_{\err, \elim}\sim D}\left( - \hat{\theta}^{\elim}_{\hatR}(\pi^*) \le \hatxi_k^{\elim}(\delta) \quad \forall k \in \mathbb{N} \right) \ge 1- \delta
\end{equation}

\begin{algorithm}[ht]
  \caption{$\CBElimErrorEstimator$}
  \label{alg:CBElimErrorEstimator}
  \textbf{Inputs:} $S_{\est}, S_{\err, \elim},R_{\elim},\hatpi_{\elim}, \delta$
 \begin{algorithmic}[1] 
  \STATE For $e^{\elim}_\pi:= \theta^{\elim}(\pi) - \hat{\theta}^{\elim}_{\hatR}(\pi) $, $\hatu_{\elim}$ defined in \eqref{eq:error-estimation-u-for-policy-error} satisfies \Cref{ass:specific-instance-error-estimate}.
  \begin{equation}
  \label{eq:error-estimation-u-for-policy-error}
      \hatu_{\elim}(\pi, S_{\err}, \delta) := \sqrt{\frac{\log(1/\delta)}{|S_\err|}}\left(\sqrt{V(p_\err, \hatpi_{\elim}) } +  \sqrt{\alpha_\err}\right) + \hat{\theta}^{\elim}_{\err}(\pi) - \hat{\theta}^{\elim}_{\hatR}(\pi),
  \end{equation}
  \STATE $\Pi_0^{\elim} := \{\pi \in \Pi \cup \hatpi_{\elim}\}$. $\hatxi^{\elim}_0(\delta) := \max_{\pi \in \Pi^{\elim}_0} \hatu_{\elim}(\pi, S_{\err}, \delta)$
  \STATE Now, for the localization step, we set $k = 0$ and
  \REPEAT
    \STATE {$k \leftarrow k+1$}
    \STATE {$\Pi^{\elim}_k \leftarrow \{\pi \in \Pi^{\elim}_{k-1}| - \hat{\theta}^{\elim}_{\hatR}(\pi) \le \hatxi^{\elim}_{k-1}(\delta)\}$}
    \STATE {$\hatxi^{\elim}_k(\delta) \leftarrow \max_{\pi \in \Pi^{\elim}_k}\hatu_{\elim}(\pi, S_{\err, \elim}, \delta)$}
  \UNTIL{$\hatxi_{k}^{\elim}(\delta) \ge \hatxi^{\elim}_{k-1}$}
  \STATE $\hatU_{\elim} \leftarrow \hatxi_{k}^{\elim}(\delta)$.
  \STATE Return $\hatU_\elim$.
  \end{algorithmic}
\end{algorithm}

\paragraph{Bounding error used in Conformal Arm Sets \eqref{eq:reward-based-error-estimation}} We now follow a similar procedure to construct a bound $\hatU_\con$ to satisfy \eqref{eq:reward-based-error-estimation}. 
In this case, our estimand is $\theta^{\con}(\pi) = R(\pi) - R(\hatpi_{\con})$, which again represents the sub-optimality of policy $\hatpi_{\con}$ with respect to a particular candidate policy $\pi$. Our goal will be to upper bound $-\theta^{\con}_{\hatf}(\pi^*)$, where
\[\theta^{\con}_{\hatf}(\pi) = R_{\hatf}(\pi) - R_{\hatf}(\hatpi_{\con}) = \mathbb{E}_{x \sim D_X}[\hatf(x, \pi(x)) - \hatf(x, \hatpi_{\con}(x))]\]
However, $\theta^{\con}_{\hatf}$ is not a directly estimable quantity (as $\theta^{\elim}_{\hatR}(\pi)$ is in \eqref{eq:cb_elim_ee-estimate_error}), as it depends on the distribution over contexts $X$. Thus, we use an estimate based $S_{\err, \con}$ for our estimate instead
\begin{equation}
\label{eq:con_estimate_error}
    \hat{\theta}^{\con}_{\hatf, \err_X}(\pi) = \mathbb{E}_{x \sim S_{\err, \con}}[\hatf(x, \pi(x)) - \hatf(x, \hatpi_{\con}(x))], \quad e^{\con}_{\pi} := \theta^{\con}(\pi) - \hat{\theta}^{\con}_{\hatf, \err_X}(\pi) 
\end{equation}
Again, as $\theta^{\con}(\pi^*) = R(\pi^*) - R(\hatpi_{\con})\ge 0$, we know that $e_{\con, \pi^*} \ge -\hat{\theta}^{\con}_{\hatf, \err_X}(\pi^*)$. As in \eqref{eq:hatu_elim_cb}, we construct $\hatu_{\con}(\pi, S_{\err, \con}, \delta)$.
\begin{align}\hatu_{\con}(\pi, S_{\err, \con}, \delta)&:= \sqrt{\frac{\log(1/\delta)}{|S_{\err, \con}|}}\left(\sqrt{V(p_\err, \hatpi_{\con}) } +  \sqrt{\alpha_\err}\right) + \hat{\theta}^{\con}_{\err}(\pi) - \hat{\theta}^{\con}_{\hatf, \err_X}(\pi), \label{eq:hatu_con_cb}
\end{align}
We then localize over the classes
\[\Pi^{\con}_k := \left\{\pi \in \Pi^{\con}_{k-1}|- \hat{\theta}^{\con}_{\hatf, \err_X}(\pi) \le \hatxi_{k}^{\con}(\delta) \right\}, \text{ where } \hatxi_k^{\con}(\delta) := \max_{\pi \in \Pi^{\con}_{k-1}} \hatu_{\con}(\pi, S_{\err, \con}, \delta) \]
By \Cref{cor:localize}, we have \eqref{eq:con_error} holds, this time we set the probability to be $1- \delta/2$.
\begin{equation}
\label{eq:con_error}
    \Pr_{S_{\err, \con}\sim D}\left( - \hat{\theta}^{\con}_{\hatf, \err_X}(\pi^*) \le \hatxi_k^{\con}(\delta/2) \quad \forall k \in \mathbb{N} \right) \ge 1- \delta/2
\end{equation}
Finally, to bound our true quantity of interest, we use Hoeffding's inequality again, 
\begin{equation}
    \Pr_{S_{\err, \con}\sim D}\left( - \theta^{\con}_{\hatf}(\pi^*) \le  - \hat{\theta}^{\con}_{\hatf, \err_X}(\pi^*) + \sqrt{\frac{2 \log (2/\delta)}{|S_{\err, \con}|}} \right) \ge 1- \delta/2
\end{equation}
See details in \Cref{app:ee}.

\begin{algorithm}[ht]
  \caption{$\CBConErrorEstimator$}
  \label{alg:CBConErrorEstimator}
  \textbf{Inputs:} $S_{\est}, S_{\err},\hat{f},\hatpi_{\con}, \delta$
 \begin{algorithmic}[1] 
  \STATE For $e^{\con}_\pi:= \theta^{\con}(\pi) - \hat{\theta}^{\con}_{\hat{f}, \err_X}(\pi)$, $\hatu_{\con}$ defined in \eqref{eq:error-estimation-u-for-regression-error} satisfies \Cref{ass:specific-instance-error-estimate}.
  \begin{equation}
  \label{eq:error-estimation-u-for-regression-error}
      \hatu_{\con}(\pi, S_{\err}, \delta):= \sqrt{\frac{\log(1/\delta)}{|S_\err|}}\left(\sqrt{V(p_\err, \hatpi_{\con}) } +  \sqrt{V(p_\err, \pi)}\right) + \hat{\theta}^{\con}_{\err}(\pi) -\hat{\theta}^{\con}_{\hat{f}, \err_X}(\pi),
  \end{equation}
  \STATE $\Pi^{\con}_0 := \{\pi \in \Pi \cup \hatpi_{\con}\}$. $\hatxi^{\con}_0(\delta) := \max_{\pi \in \Pi^{\con}_0} \hatu_{\con}(\pi, S_{\err, \con}, \delta/3)$
  \STATE Now, for the localization step, we set $k = 0$ and
  \REPEAT
    \STATE {$k \leftarrow k+1$}
    \STATE {$\Pi^{\con}_k \leftarrow \{\pi \in \Pi^{\con}_{k-1}| -\hat{\theta}^{\con}_{\hat{f}, \err_X}(\pi) \le \hatxi^{\con}_{k-1}(\delta)\}$}
    \STATE {$\hatxi^{\con}_k(\delta) \leftarrow \max_{\pi \in \Pi_k}\hatu_{\con}(\pi, S_{\err, \con}, \delta)$}
  \UNTIL{$\hatxi^{\con}_{k}(\delta) \ge \hatxi^{\con}_{k-1}(\delta)$}
  \STATE $\hatU_{\con} \leftarrow \hatxi^{\con}_{k}(\delta/2) $.
  \STATE Return $\hatU_\con + \sqrt{\frac{2\log(2/\delta)}{|S_{\err, \con}|}}$.
  \end{algorithmic}
\end{algorithm}

\begin{corollary}
\label{cor:base_cb_ee_result}
Given $S_\err, S_\est, R_{\elim}, \hatpi_{\elim}, \delta$, \eqref{eq:ee_policy_bound} holds.
    \begin{equation}
        \Pr_{S_{\err, \elim} \sim \calD}\left(\hat{R}_{\elim}(\hatpi_{\elim}) - \hat{R}_{\elim} (\pi^*) \le \hatxi^{\elim}_k (\delta)\quad \forall k \in \mathbb{N}\right) \ge 1- \delta, \label{eq:ee_policy_bound}
    \end{equation}
    Independently, given $S_\err, S_\est, \hat{f}, \hatpi_{\con}, \delta$, \eqref{eq:ee_reward_bound} holds.
    \begin{equation}
         \Pr_{S_{\err, \con} \sim \calD}\left(R_{\hat{f}}(\hatpi_{\con}) - R_{\hat{f}}(\pi^*) \le \hatxi^{\con}_k(\delta/2) + \sqrt{\frac{2\log(2/\delta)}{|S_{\err, \con}|}} \quad \forall k \in \mathbb{N}\right) \ge 1- \delta, \label{eq:ee_reward_bound}
    \end{equation}
Hence, \eqref{eq:policy-based-error-estimation} and \eqref{eq:reward-based-error-estimation} hold.
\end{corollary}

\subsection{Contextual Arm Elimination}
\label{sec:ArmEliminator}
In this section we construct a simple Arm Eliminator which uses an upper confidence bound (UCB) style argument to guess which arms to include, then checks that no arms which perform well enough under $\hatR_{\elim}$ have been excluded. If there exist such arms, $\ArmEliminator$ includes more arms.
\begin{algorithm}[ht]
  \caption{$\ArmEliminator$}
  \label{alg:ArmEliminator}
  \textbf{Inputs:} $\hatR_{\elim},\hat{\pi}_{\elim},\hatU_{\elim},S$.
  \begin{algorithmic}[1] 
  \STATE Estimate $\hatf_{\mean},\hatf_{\width}\leftarrow \BlackBoxCIEstimator(S)$
  \STATE $\ValidEliminator\leftarrow\textbf{False}$, $\IntervalMultiplier\leftarrow 1$. 
  \WHILE{\textbf{not} \ValidEliminator }
    \STATE Set $\hatg(x)\leftarrow \{a|\hatf_{\mean}(x,a) + \IntervalMultiplier*\hatf_{\width}(x,a)  \geq \max_a (\hatf_{\mean}(x,a) - \IntervalMultiplier*\hatf_{\width}(x,a))   \}$ for all $x$.
    \STATE Get $\ValidEliminator\leftarrow\ValidateEliminator(\hatR_{\elim},\hat{\pi}_{\elim},\hatU',S)$.
    \STATE $\IntervalMultiplier\leftarrow 2*\IntervalMultiplier$. 
    \ENDWHILE
    \STATE Return $\hatg$.
  \end{algorithmic}
\end{algorithm}
Where $\ValidateEliminator$ is designed to output \eqref{eq:ideal-eliminator-validator}, which we approximate with \eqref{eq:real-eliminator-validator}. 
\begin{align}
    &\mathbf{1}\Bigg(\min_{\{\pi \in \Tilde{\Pi} | \pi(x) \notin g(x) \forall x \} } \hatR_{\elim}(\pi) - \hatR_{\elim}(\pi^*)> \hatU_{\elim} \Bigg) \label{eq:ideal-eliminator-validator}\\
    & \approx \mathbf{1}\Bigg(\min_{\{\pi \in \Tilde{\Pi} | \pi(x) \notin g(x) \forall x \in S \} } \hatR_{\elim}(\pi) - \hatR_{\elim}(\pi^*)> \hatU_{\elim} \Bigg) =\ValidateEliminator(\hatR_{\est},\hat{\pi}_{\est},\hatU',S). \label{eq:real-eliminator-validator}
\end{align}
\textbf{Informal guarantee:} It is easy to verify that under the ideal $\ValidateEliminator$ \eqref{eq:ideal-eliminator-validator}, from \Cref{cor:base_cb_ee_result}, we have \eqref{eq:ideal-arm-eliminator-guarantee} holds. Moreover, if the real $\ValidateEliminator$ \eqref{eq:real-eliminator-validator} uses a holdout set $S'$ instead of $S$, by adding a $O(\sqrt{\log(|S'|)/n})$ term to the L.H.S. of the inequality in \eqref{eq:real-eliminator-validator} -- we can ensure \eqref{eq:ideal-arm-eliminator-guarantee} holds with high-probability for $(x,r)\sim D$ -- for notational simplicity, we do not introduce this additional holdout set.

%




\subsection{Details for Contextual Bandit Pipeline}
In this Appendix we prove \Cref{thm:cbpipelinemain} for the Contextual Bandit pipeline and formalize the error estimation argument in the Contextual Bandit setting.
\subsubsection{Proof of Theorem 2}

We begin relating conformal arm sets to classical notions in conformal prediction (\cite{vovk_conformal_2005}).
\begin{lemma}[Conformal Uncertainty]
\label{lem:conformal-uncertainty} 
Suppose $R_{\hatf}(\hatpi_{\con}) - R_{\hatf}(\pi^*)\leq \hatU_{\con}$, we then have:
\begin{equation}
    \Pr_{x\sim D_{\calX}}(\pi^*(x)\in C(x,\zeta)) \geq 1 - \zeta,\;\;\; \forall \zeta\in(0,1].
\end{equation}
\end{lemma}

\begin{proof}
Suppose $R_{\hatf}(\hatpi_{\con}) - R_{\hatf}(\pi^*)\leq \hatU_{\con}$, we then have \eqref{eq:confidence_set_CAS_EE} holds.
\begin{equation}
\label{eq:confidence_set_CAS_EE}
\begin{aligned}
        &\Pr_{x\sim D_{\calX}}(\pi^*(x)\notin C(x,\zeta))\\
        &\stackrel{(i)}{=} \Pr_{x\sim D_{\calX}}\Big( \hatf (x, \hatpi_{\con}(x)) - \hatf (x,\pi^*(x))> \frac{\hatU_{\con}}{\zeta}\Big)\\
        &\stackrel{(ii)}{\leq} \frac{ \E_{x\sim D_{\calX}}[\hatf (x, \hatpi_{\con}(x)) - \hatf (x,\pi^*(x))]}{\hatU_{\con}/\zeta} \stackrel{(iii)}{=}  \frac{ R_{\hatf}(\hatpi_{\con}) - R_{\hatf}(\pi^*)}{\hatU_{\con}/\zeta} \stackrel{(iv)}{\leq} \zeta.
\end{aligned}
\end{equation}
Here (i) follows from \eqref{eq:CAS-based-on-error-estimation}, (ii) follows from Markov's inequality, (iii) follows by definition of $R_{\hatf}$ \eqref{eq:def-reward-and-cover}, and (iv) follows from $R_{\hatf}(\hatpi_{\con}) - R_{\hatf}(\pi^*)\leq \hatU_{\con}$.
\end{proof}
Now, we formalize the intuition that if an arm $a$ is "likely to be a good arm for context $x$" (either by belonging to both $\hatg(x)$ and $C(x, \beta_{\max}/\eta)$, or just to $\hatg(x)$), then the learner has a positive probability of pulling arm $a$ under exploration policy $p$.
\begin{lemma}
\label{lemma:lower_bound_p_error_est}
    We have \eqref{eq:pm_lower_bound} holds.
    \begin{equation}
        \label{eq:pm_lower_bound}
        \begin{aligned}
        p(a|x)&\geq \begin{cases}
			\frac{1-\propThreshold}{|C(x, \propThreshold/\eta)|} + \frac{\propThreshold}{|\hatg(x)|}, & \text{if $a\in C(x, \propThreshold/\eta)$}\\
            \frac{\eta}{|\hatg(x)|} \frac{\hatU_{\con}}{\hatf(x,\hatpi_{\con}(x))-\hatf(x,a)}, & \text{if $a\notin C(x, \propThreshold/\eta)$ and $a\in\hatg(x)$}\\
            0, & \text{otherwise.}
		 \end{cases}\\
            &\geq \begin{cases}
			\frac{1}{|\hatg(x)|}, & \text{if $a\in C(x, \propThreshold/\eta)$}\\
            \frac{\eta\hatU_{\con}}{|\hatg(x)|}, & \text{if $a\notin C(x, \propThreshold/\eta)$ and $a\in\hatg(x)$}\\
            0, & \text{otherwise.}
		 \end{cases}
        \end{aligned}
    \end{equation}
\end{lemma}
\begin{proof}
Recall that $\Unif_{\zeta}(\cdot|x)$ denotes the uniform distribution over arms in $C(x,\zeta)$ and $p$ is given by \eqref{eq:define_pm_genRAPR_restated}.  
\begin{equation}
\label{eq:define_pm_genRAPR_restated}
\begin{aligned}
    &p (a|x) = (1-\beta_{\max})\Unif_{\beta_{\max}/\eta}(a|x)+ \int_{0}^{\beta_{\max}}\Unif_{\beta/\eta}(a|x)\mbox{d}\beta.
\end{aligned}
\end{equation}
Clearly if $a\notin \hatg(x)$, we have $p(a|x)=0$. The rest of our analysis to lower bound $p(a|x)$ is divided into two cases based on whether $a$ lies in $C(x, \propThreshold/\eta)$.

\textbf{Case 1 ($a\in C(x, \propThreshold/ \eta)$).} Note that $C(x, \propThreshold/ \eta)\subseteq C(x, \beta/ \eta) \subseteq \A$ for all $\beta\in[0,\propThreshold]$. Hence, $a\in C(x, \beta/\eta)$ for all $\beta\in[0,\propThreshold]$. Therefore, in this case, $p (a|x)\geq \frac{1-\propThreshold}{|C(x, \propThreshold/\eta)|} + \frac{\propThreshold}{|\hatg(x)|}\geq \frac{1}{|\hatg(x)|}$.

\textbf{Case 2 ($a\notin C(x, \propThreshold/\eta)$ and $a\in\hatg(x)$).} For this case, the proof follows from \eqref{eq:lower_bound_term2_in_pm}.

\begin{equation}
\label{eq:lower_bound_term2_in_pm}
    \begin{aligned}
        p(a|x) \geq & \int_0^{\propThreshold}\frac{I[a\in C(x,\beta/\eta) ]}{|C(x,\beta/\eta)|}\mbox{d}\beta\\
       \stackrel{(i)}{\geq}  & \frac{1}{|\hatg(x)|} \int_0^{\propThreshold}I[a\in C(x,\beta/\eta)]\mbox{d}\beta = \frac{1}{|\hatg(x)|} \int_0^{\propThreshold}I[a\in C(x,\beta/\eta)]\mbox{d}\beta\\
       \stackrel{(ii)}{\geq}  & \frac{I[a\notin C(x,\propThreshold/\eta)]}{|\hatg(x)|} \int_0^{\propThreshold}I[a\in C(x,\beta/\eta) ]\mbox{d}\beta\\
       \stackrel{(iii)}{=}  & \frac{I[a\notin C(x,\propThreshold/\eta)]}{|\hatg(x)|} \int_0^{1}I[a\in C(x,\beta/\eta) ]\mbox{d}\beta\\
      \stackrel{(iv)}{=} & \frac{I[a\notin C(x,\propThreshold/\eta)]}{|\hatg(x)|} \int_0^{1} I\bigg[\hatf (x, \hatpi_{\con}(x)) - \hatf (x,a)\leq \frac{\eta\hatU_{\con} }{\beta}\bigg]\mbox{d}\beta\\
      = & \frac{I[a\notin C(x,\propThreshold/\eta)]}{|\hatg(x)|} \int_0^{1} I\bigg[\beta\leq \frac{\eta\hatU_{\con}}{\hatf (x, \hatpi_{\con}(x)) - \hatf (x,a)}\bigg]\mbox{d}\beta\\
      \stackrel{(v)}{=} & \frac{I[a\notin C(x,\propThreshold/\eta)]}{|\hatg(x)|}   \frac{\eta\hatU_{\con}}{\hatf (x, \hatpi_{\con}(x)) - \hatf (x,a)}\\
      \stackrel{(vi)}{\geq} & \frac{\eta\hatU_{\con} I[a\notin C(x,\propThreshold/\eta)]}{|\hatg(x)|} 
    \end{aligned}
\end{equation}
where (i) is because the size of a set's intersection with $\hatg(x)$ can be no larger than the size of $\hatg(x)$; (ii) follows from $I(a\notin C(x,\propThreshold/\eta))\leq 1$; (iii) follows from the fact that if $a\notin C(x,\propThreshold/\eta)$ then $a\notin C_{m}(x,\beta/\eta)$ for all $\beta\geq \propThreshold$; (iv) follows from \eqref{eq:CAS-based-on-error-estimation}; (v) follows from the fact that $ \frac{\eta\hatU_{\con}}{\hatf (x, \hatpi_{\con}(x)) - \hatf (x,a)}\leq 1$ when $a\notin C(x,\propThreshold/\eta)$; and (vi) follows from the fact that $\hatf:\calX\times\calA\rightarrow [0,1]$.
\end{proof}

Now, in \Cref{lemma:bound_v_error}, we use the bounds derived in \Cref{lemma:lower_bound_p_error_est} to establish bounds on the optimal cover $V(p, \pi^*)$.
 
\begin{lemma}
\label{lemma:bound_v_error}
Suppose $R_{\hatf}(\hatpi_{\con}) - R_{\hatf}(\pi^*) \leq \hatU_\con$ and $\hatR_{\elim}(\hat{\pi}_{\elim}) - \hatR_{\elim}(\pi^*)\leq \hatU_{\elim}$. Further suppose \eqref{eq:ideal-arm-eliminator-guarantee} holds, therefore $\pi^*(x)\in\hatg(x)$ for all $x$. We then have \eqref{eq:v_upper_bound_opt_policy_elim} holds.
\begin{equation}
    \label{eq:v_upper_bound_opt_policy_elim}
    \begin{aligned}
    V(p, \pi^*)&\leq \E_{x\sim D_{\calX}}\Bigg[\frac{|C(x, \propThreshold/\eta)|}{1-\propThreshold + \propThreshold \frac{|C(x, \propThreshold/\eta)|}{|\hatg(x)|} } + \frac{|\hatg(x)|}{\eta}\frac{(\hatf (x, \hatpi_{\con}(x)) - \hatf (x,\pi^*(x)))}{\hatU_{\con}}\Bigg]\\
    &\leq \E_{x\sim D_{\calX}}\Bigg[\frac{|C(x, \propThreshold/\eta)|}{1-\propThreshold + \propThreshold \frac{|C(x, \propThreshold/\eta)|}{|\hatg(x)|} }\Bigg] + \frac{\max_x |\hatg(x)|}{\eta}.
    \end{aligned}
\end{equation}
\end{lemma}

\begin{proof}
From \Cref{lemma:lower_bound_p_error_est}, we have \eqref{eq:pm_lowerbounds} holds. 
\begin{equation}
\label{eq:pm_lowerbounds}
    \begin{aligned}
        & \frac{I(\pi^*(x)\in C(x, \propThreshold/\eta))}{p(\pi^*(x)|x)}\leq  \frac{1}{\frac{1-\propThreshold}{|C(x, \propThreshold/\eta)|} + \frac{\propThreshold}{|\hatg(x)|}}\leq \frac{|C(x, \propThreshold/\eta)|}{1-\propThreshold + \propThreshold \frac{|C(x, \propThreshold/\eta)|}{|\hatg(x)|} }, \\
        \mbox{and}\quad & \frac{I(\pi^*(x)\notin C(x, \propThreshold/ \eta))}{p(\pi^*(x)|x)}\leq \frac{|\hatg(x)|}{\eta} \frac{\hatf(x,\hatpi_{\con}(x))-\hatf(x,\pi^*(x))}{\hatU_{\con}}.
    \end{aligned}
\end{equation}
We now bound the optimal cover as follows,
\begin{equation}
\begin{aligned}
    &V(p,\pi^*)=\E_{x\sim D_{\calX}}\bigg[\frac{1}{p(\pi^*(x)|x)}\bigg]\\ 
    = &\E_{x\sim D_{\calX}}\bigg[\frac{I[\pi^*(x)\in C(x,\propThreshold/ \eta)] + I[\pi^*(x)\notin C(x,\propThreshold/\eta)]}{p(\pi^*(x)|x)}\bigg]\\
    \stackrel{(i)}{\leq}& \E_{x\sim D_{\calX}}\Bigg[\frac{|C(x, \propThreshold/\eta)|}{1-\propThreshold + \propThreshold \frac{|C(x, \propThreshold/\eta)|}{|\hatg(x)|} } + \frac{|\hatg(x)|}{\eta}\frac{(\hatf (x, \hatpi_{\con}(x)) - \hatf (x,\pi^*(x)))}{\hatU_{\con}}\Bigg]\\
    \stackrel{(ii)}{\leq} &\E_{x\sim D_{\calX}}\Bigg[\frac{|C(x, \propThreshold/\eta)|}{1-\propThreshold + \propThreshold \frac{|C(x, \propThreshold/\eta)|}{|\hatg(x)|} }\Bigg] + \frac{\max_x|\hatg(x)|}{\eta}. 
\end{aligned}
\end{equation}
Here (i) follows from from \eqref{eq:pm_lowerbounds} and (ii) follows from $R_{\hatf}(\hatpi_{\con}) - R_{\hatf}(\pi^*)\leq \hatU_{\con}$.
\end{proof}
Finally, we combine results from the Lemmas above to prove \Cref{thm:cbpipelinemain}.
\cbpipelinemain*
\begin{proof}
Since the bounds in $S$ hold with probability at least  $1-\deltaprior$ and under this event \eqref{eq:policy-based-error-estimation} holds, we have: the bounds in $S$, $R_{\hatf}(\hatpi_{\con}) - R_{\hatf}(\pi^*)\leq \hatU_{\con}$, and $\hatR_{\elim}(\hat{\pi}_{\elim}) - \hatR_{\elim}(\pi^*)\leq \hatU_{\elim}$ hold with probability at least $1-\deltaprior-2\delta/3$. Hence, by the union  bound, it is sufficient to prove \eqref{eq:pipeline-guarantees} with probability at least $1 - \delta/3$ under these events. Therefore moving forward, we assume $R_{\hatf}(\hatpi_{\con}) - R_{\hatf}(\pi^*)\leq \hatU_{\con}$ and $\hatR_{\elim}(\hat{\pi}_{\elim}) - \hatR_{\elim}(\pi^*)\leq \hatU_{\elim}$ hold.

Now note that $\hatg$ is computed in a way that satisfies \eqref{eq:ideal-arm-eliminator-guarantee}. Since $\hatR_{\elim}(\hat{\pi}_{\\elim}) - \hatR_{\elim}(\pi^*)\leq \hatU_{\elim}$, we therefore have $\hatg(x)\in\pi^*(x)$ for all $x$. Hence from $R_{\hatf}(\hatpi_{\con}) - R_{\hatf}(\pi^*)\leq \hatU_{\con}$, $\hatg(x)\in\pi^*(x)$ for all $x$, and \Cref{lem:conformal-uncertainty,lemma:lower_bound_p_error_est,lemma:bound_v_error}, we have \eqref{eq:first_pipeline-guarantees} holds.
\begin{equation}
    \label{eq:first_pipeline-guarantees}
        \begin{aligned}
            &\pi^*(x)\in \hatg(x),\forall x;\;\;\;  \max_{x}\frac{1}{p(\pi^*(x)|x)} \leq  \frac{\max_x |\hatg(x)|}{\eta\hatU_{\con}};\\ 
            &V(p,\pi^*)\leq  \E_{x\sim \calD_x}\Bigg[\frac{|C(x, \propThreshold/\eta)|}{1-\propThreshold + \propThreshold \frac{|C(x, \propThreshold/\eta)|}{|\hatg(x)|} }\Bigg] + \frac{\max_x |\hatg(x)|}{\eta} 
        \end{aligned}
    \end{equation}
    Finally, by Hoeffding's inequality, with probability at least $1- \delta/3$, we have
    \begin{align*}
         V(p,\pi^*)&\leq\E_{x\sim \calD_x}\Bigg[\frac{|C(x, \propThreshold/\eta)|}{1-\propThreshold + \propThreshold \frac{|C(x, \propThreshold/\eta)|}{|\hatg(x)|} }\Bigg] + \frac{\max_x |\hatg(x)|}{\eta}\\
         &\le  \E_{x\sim S_{\err, B}}\Bigg[\frac{|C(x, \propThreshold/\eta)|}{1-\propThreshold + \propThreshold \frac{|C(x, \propThreshold/\eta)|}{|\hatg(x)|} }\Bigg] + \sqrt{\frac{\max_x | \hatg(x)|\log(3/\delta)}{\beta_{\max}|S_{\err, B}|}} + \frac{\max_x |\hatg(x)|}{\eta}.
    \end{align*}
\end{proof}

\subsubsection{Proof of Error Estimation for Contextual Bandits}
\label{app:ee}
In this section, we present the steps required to apply the Error Estimation method to the Contextual Bandit setting, and ultimately prove \Cref{cor:base_cb_ee_result}.

\begin{proof}
Recall that we have two subroutines:
\begin{itemize}
    \item $\CBElimErrorEstimator$ constructs $\hatU_{\elim}$ as a bound on the error in the Policy Evaluation step, such that \eqref{eq:policy-based-error-estimation} holds:
    \[\Pr_{S_{\err, \elim} \sim \calD}\left(\hatR_{\elim}(\hatpi_{\elim}) - \hatR_{\elim}(\pi^*) \le \hatU_{\elim}\right) \ge 1 - \delta/3\]
    \item $\CBConErrorEstimator$ constructs $\hatU_{\con}$ as a bound on the error in the Reward model step, such that \eqref{eq:reward-based-error-estimation} holds:
    \[\Pr_{S_{\err, \con} \sim \calD}\Big(R_{\hatf}(\hatpi_{\con}) - R_{\hatf}(\pi^*) \leq \hatU_\con \Big)\geq 1-\delta/3\]
\end{itemize}

Recall the Key Error Estimation Quantities for Arm Elimination and Conformal Arm sets in \Cref{tab:ee_cb}.

\textbf{Step 1: Bounding the error in Policy Evaluation.} Recall that we want to bound $-\hat{\theta}^{\elim}_{\hatR}(\pi^*)$, and since $\theta^{\elim}(\pi^*) = R(\pi^*) - R(\hat{\pi}_{\elim})\ge 0$ (recall we define $\pi^* = \arg\max_{\pi \in \Pi} R(\pi)$, and construct $\Pi$ to include $\hat{\pi}_{\elim}$), it suffices to find an upper bound for $e^{\elim}_{\pi^*}$. Using the error estimation heuristic, we introduce $\hat{\theta}^{\con}_{\err}(\pi)$ as in \Cref{tab:ee_cb}, and write
\begin{align*}
    e^{\elim}_{\pi} &=  \theta^{\elim}(\pi) - \hat{\theta}^{\elim}_{\hatR}(\pi)  + \hat{\theta}^{\elim}_{\err}(\pi) - \hat{\theta}^{\elim}_{\err}(\pi) \\
    &= \underbrace{\hat{\theta}^{\elim}_{\err}(\pi) - \hat{\theta}^{\elim}_{\hatR}(\pi)}_{A} + \underbrace{\theta^{\elim}(\pi)-\hat{\theta}^{\elim}_{\err}(\pi)}_{B}
\end{align*}
\begin{enumerate}
        \item We can find $A$ directly as both terms are estimable. Note that we estimate the first term using inverse propensity scores. $A = \mathbb{E}_{(a,x) \sim S_\err}\left[\frac{ \mathbf{1}_{a = a'} r(a',x)(\pi(a'|x) - \hatpi_{\elim}(a'|x)}{p_\err(a'|x)}\right] - \hat{\theta}^{\elim}_{\hatR}(\pi)$.
        \item For $B$, note that this term corresponds to $e^{\elim}_{\err, \pi}$ from the general framework introduce in \Cref{sec:examples}. We now construct a bound that satisfies \Cref{ass:specific-instance-error-estimate}. In particular, by Freedman's inequality \citep{dudik2011efficient}, we have for any $\pi \in \Tilde{\Pi}$ 
        \begin{equation}
            B \le \sqrt{\frac{\log(1/\delta)}{n}} \left( \sqrt{V(p_{\err}, \pi)} + \sqrt{V(p_{\err}, \hatpi_{\elim})}\right)
        \end{equation} with probability at least $1-\delta$ 
    \end{enumerate}
     Combining these terms, we have 
\[e^{\elim}_{\pi} \le \sqrt{\frac{\log(1/\delta)}{n}} \left(\sqrt{V(p_\err, \pi)} +  \sqrt{V(p_\err, \hatpi_{\elim})}\right)  + \hat{\theta}^{\elim}_{\err}(\pi) - \hat{\theta}^{\elim}_{\hatR}(\pi)\] 
Now we define 
\begin{equation}
    \hatu_{\elim}(\pi, S_\err, \delta)  := \sqrt{\frac{\log(1/\delta)}{n}} \left(\sqrt{\alpha_{\err}} +  \sqrt{V(p_\err, \hatpi_{\elim})}\right)  + \hat{\theta}^{\elim}_{\err}(\pi) - \hat{\theta}^{\elim}_{\hatR}(\pi)
\end{equation}
Note that $\hatu_{\elim}(\pi, S_\err, \delta)$ satisfies \Cref{ass:specific-instance-error-estimate} rather than \Cref{ass:base-error-estimate}, as we use $\alpha_\err \ge V(p_\err, \pi^*)$ for the specific instance $\pi^*$. Thus, we have the initial bound
\begin{equation}
\label{eq:policy_error_bound_firstpass}
   e_{\pi^*}^{\elim}\le \hatu_{\elim}(\pi^*, S_\err, \delta) \le \max_{\pi \in \Pi} \left( \hatu_{\elim}(\pi, S_\err, \delta)\right)
\end{equation}

\textbf{Step 2: Improving the error bound in Policy Evaluation using localization.}
Now, given $\theta^{\con}(\pi^*) \ge 0$, we have satisfied \Cref{ass:h*bound}, and can thus apply \Cref{cor:localize}. Let $\Pi^{\elim}_0 = \Tilde{\Pi}$, and define
\[\hatxi^{\elim}_k(\delta):= \max_{\pi \in \Pi^{\elim}_{k-1}}\hatu_{\elim}(\pi, S_\err, \delta),\;\;\; \Pi^{\elim}_k :=  \left\{\pi \in \Pi^{\elim}_{k-1}|-\hat{\theta}^{\elim}_{\hatR}(\pi)\le \xi_k\right\},\]
Recalling that $-\hat{\theta}^{\elim}_{\hatR}(\pi^*) \le e^{\elim}_{\pi^*}$, this gives us the final bound
\[\Pr_{S_{\err, \elim} \sim \calD}\left( -\hat{\theta}^{\elim}_{\hatR}(\pi^*) \le \hatxi^{\elim}_k(\delta) \forall k \right) \ge 1- \delta.\] \\
\textbf{Step 3: Bounding the error in the reward model.}
 For the reward model, recall that we are interested in bounding $-\theta^{\con}_{\hat{f}}(\pi^*)$, where
  \[\theta^{\con}_{\hat{f}}(\pi)  = R_{\hat{f}}(\hatpi_{\con}) - R_{\hat{f}}(\pi) = \mathbb{E}_{x \sim \calD}[\hatf(x, \hat{\pi}_{\con}(x) - \hatf(x, \pi(x))].\]
  However, unlike the quantity in arm elimination, this is not directly computable as it an average over an unknown distribution of contexts. Thus we will instead use the estimate 
  \[\hat{\theta}^{\con}_{\hat{f}, \err_X}(\pi) = \mathbb{E}_{(x,a,r) \sim S_{\err}}[R(\pi) - R(\hat{\pi}_{\con})],\] and use error estimation to upper bound the maximum error, and then use a single error estimate to bound the difference between this estimate and our true estimate of interest. This case is now almost identical to the arm elimination case. We similarly define $e^{\con}_\pi = \theta^{\con}(\pi) - \hat{\theta}^{\con}_{\hat{f}, \err_X}(\pi)$. Since $\theta^{\con}(\pi^*) ( = R(\pi) - R(\hat{\pi}_{\con}))\ge 0$, it suffices to find an upper bound for $e^{\con}_{\pi^*}$ to bound $- \hat{\theta}^{\con}_{\hat{f}, \err_X}(\pi^*)$. 

\begin{align*}
    e_{\pi} &= \theta^{\con}(\pi) - \hat{\theta}^{\con}_{\hatf, \err_X}(\pi) + \hat{\theta}^{\con}_{\err}(\pi) - \hat{\theta}^{\con}_{\err}(\pi)\\
    &= \underbrace{\hat{\theta}^{\con}_{\err}(\pi) - \hat{\theta}^{\con}_{\hatf, \err_X}(\pi)}_{A} + \underbrace{\theta^{\con}(\pi) - \hat{\theta}^{\con}_{\err}(\pi)}_{B} 
    \end{align*}
    \begin{enumerate}
        \item Estimating $\hat{\theta}^{\con}_{\hat{f}, \err_X}(\pi)$ directly and using an unbiased estimate to evaluate $\hat{\theta}^{\con}_{\err}(\pi)$. 
            \begin{align*}
        \hat{\theta}^{\con}_{\err}(\pi) &= \hat{R}^{\err}(\pi) - \hat{R}^{\err}(\hatpi_{\con})\\
        &= \mathbb{E}_{(a,x) \sim S_\err}\left[\sum_{a' \in \calA} \frac{ \mathbf{1}_{a = a'} r(a',x)(\pi(a'|x) - \hatpi_{\con}(a'|x)}{p_\err(a'|x)}\right]
        \end{align*}
        So
        \[A = \mathbb{E}_{(a,x) \sim S_\err}\left[\sum_{a' \in \calA} \frac{ \mathbf{1}_{a = a'} r(a',x)(\pi(a'|x) - \hatpi_{\con}(a'|x)}{p_\err(a'|x)}\right] - \hat{\theta}^{\con}_{\hat{f}, \err_X}(\pi)\]
    \item  Note that we can write
    \begin{align*}
        \theta^{\con}(\pi) - \hat{\theta}^{\con}_{\err}(\pi) &= R(\pi) - \hat{R}^{\err}(\pi) + \hat{R}^{\err}(\hatpi_{\text{\con}}) -R(\hatpi_{\text{\con}})
    \end{align*}
    From Freedman's inequality \citep{dudik2011efficient}, we have with probability at least $1-\delta/2$
    \[B \le \sqrt{\frac{\log(2/\delta)}{n}} \left( \sqrt{V(p_{\err}, \pi)} + \sqrt{V(p_{\err}, \hatpi_{\con})}\right)\]
    \end{enumerate}
    
    Combining the terms, we have that for any given policy $\pi$, with probability at least $1-\delta$.
\[e^{\con}_{\pi} \le \sqrt{\frac{\log(1/\delta)}{n}} \left(\sqrt{V(p_\err, \hatpi_{\con}) }+\sqrt{V(p_\err, \pi)}\right)  + \hat{\theta}^{\con}_{\err}(\pi) - \hat{\theta}^{\con}_{\hat{f}, \err_X}(\pi)\]
Now, as above, we define $\hatu_{\con}(\pi, S_\err, \delta)$ to satisfy \Cref{ass:specific-instance-error-estimate}.
\begin{equation}
    \hatu_{\con}(\pi, S_\err, \delta) := \sqrt{\frac{\log(1/\delta)}{n}} \left( \sqrt{V(p_\err, \hatpi_{\con}) }+\sqrt{\alpha_\err}\right)  + \hat{\theta}^{\con}_{\err}(\pi) - \hat{\theta}^{\con}_{\hat{f}, \err_X}(\pi)
\end{equation}
Now, since $\theta^{\con}(\pi^*) \ge 0$, we have satisfied \Cref{ass:h*bound} and can apply \Cref{cor:localize}. We note let $\Pi^{\con}_0 = \Tilde{\Pi}$, and define 
 \[\hatxi^{\con}_k(\delta/2):= \max_{\pi \in \Pi^{\con}_{k-1}}\hatu_{\con}(\pi, S_{\err, \con}, \delta/2),\;\;\; \Pi^{\con}_k :=  \left\{\pi \in \Pi^{\con}_{k-1}|-\theta^{\con}_{\hatf, \err_X}(\pi)\le \hatxi_k(\delta/2)\right\},\]
 This gives us
 \[\Pr_{S_{\err, \con} \sim \calD}\left( -\hat{\theta}^{\con}_{\hatf, \err_X}(\pi^*) \le \hatxi^{\con}_k (\delta/2) \forall k \right) \ge 1- \delta/2.\]
Finally, to account for having used the estimate $\hat{\theta}_{\hatf, \err_X}(\pi)$ rather than $\theta_{\hatf}(\pi)$, we bound the difference by recognizing that $\hatf$ is determined independently of $S_{\err}$, and we can therefore use a classic Hoeffding inequality, along with a union bound argument to find that with probability at least $1-\delta/2$,
    \[\Pr_{S_{\err, \con} \sim \calD}\left( -\theta^{\con}_{\hatf}(\pi^*) \le \hatxi^{\con}_k(\delta/2) \forall k  + \sqrt{\frac{2\log(2/\delta)}{|S_{\err, \con}|}}\right) \ge 1- \delta.\]
\end{proof}


